\documentclass[10pt,twocolumn,letterpaper]{article}
\usepackage[T1]{fontenc}

\usepackage[pagenumbers]{cvpr}

\usepackage{cuted}
\usepackage{currfile}
\usepackage{caption}
\usepackage{graphicx}
\usepackage{afterpage}
\usepackage{float}
\usepackage{amsmath,amssymb,bm}
\usepackage{booktabs}
\usepackage{multirow}
\usepackage{makecell}
\usepackage{colortbl}
\usepackage{xspace}
\usepackage{pifont}
\usepackage{enumitem}
\usepackage{tikz}
\usetikzlibrary{arrows.meta,positioning,calc,fit,backgrounds,shapes.misc,decorations.pathreplacing}
\usepackage{microtype}

\newcommand{\ours}{CoVeR\xspace}
\newcommand{\base}{LLM-Fusion\xspace}
\newcommand{\bench}{MR-Bench\xspace}
\newcommand{\benchA}{MR-Bench-A\xspace}
\newcommand{\benchH}{MR-Bench-H\xspace}
\newcommand{\shorterr}{ShortErr\xspace}

\definecolor{ctext}{HTML}{2A78D6}
\definecolor{caudio}{HTML}{EB6834}
\definecolor{cvis}{HTML}{1BAF7A}
\definecolor{cprior}{HTML}{898781}
\definecolor{cbase}{HTML}{8A8A8A}
\definecolor{cours}{HTML}{4A3AA7}
\definecolor{cbad}{HTML}{C0392B}
\definecolor{cgood}{HTML}{1E7B34}
\definecolor{oursrow}{HTML}{EEEBFA}
\definecolor{cmllm}{HTML}{6E7FA3}

\newcommand{\sg}{\operatorname{sg}}
\newcommand{\sgn}{\operatorname{sgn}}
\newcommand{\1}{\mathbb{I}}
\newcommand{\M}{\mathcal{M}}
\newcommand{\V}{\mathcal{V}}
\newcommand{\Luni}{\mathcal{L}_{\mathrm{uni}}}

\newcommand{\cmark}{\ding{51}}
\newcommand{\xmark}{\ding{55}}
\newcommand{\up}{\ensuremath{\uparrow}}
\newcommand{\dn}{\ensuremath{\downarrow}}

\newcommand{\para}[1]{\vspace{2pt}\noindent\textbf{#1}\hspace{0.5em}}

\definecolor{best}{RGB}{244,181,180}
\definecolor{second}{RGB}{249,219,184}
\definecolor{third}{RGB}{253,244,197}

\definecolor{cvprblue}{rgb}{0.21,0.49,0.74}
\usepackage[pagebackref,breaklinks,colorlinks,allcolors=cvprblue]{hyperref}
\hypersetup{pdftitle={Words Speak Louder? Exploring the Language Shortcut of LLM-Driven Incomplete Multimodal Learning},pdfauthor={Qiangchang Wang, Chong Xiao, Xiaokun Bai, Rundong He, Zhuang Qi, Xusheng Liang, Zhibin Wu, Wenhao Li}}

\definecolor{best}{RGB}{244,181,180}
\definecolor{second}{RGB}{249,219,184}
\definecolor{third}{RGB}{253,244,197}

\def\confName{CVPR}
\def\confYear{2027}

\title{Words Speak Louder? Exploring the Language Shortcut\\of LLM-Driven Incomplete Multimodal Learning}

\author{Qiangchang Wang$^{1}$\quad Chong Xiao$^{1}$\quad Xiaokun Bai$^{1}$\quad Rundong He$^{3}$\\
Zhuang Qi$^{4}$\quad Xusheng Liang$^{5,2}$\quad Zhibin Wu$^{1}$\quad Wenhao Li$^{1,2}$\\[3pt]
{\normalsize $^{1}$Shandong University\quad $^{2}$Shenzhen Loop Area Institute}\\
{\normalsize $^{3}$The Hong Kong Polytechnic University\quad $^{4}$National University of Singapore}\\
{\normalsize $^{5}$City University of Hong Kong}
}

\begin{document}
\maketitle
\begin{strip}
\par\vspace{-34pt}
\centering
\includegraphics[width=0.60\textwidth]{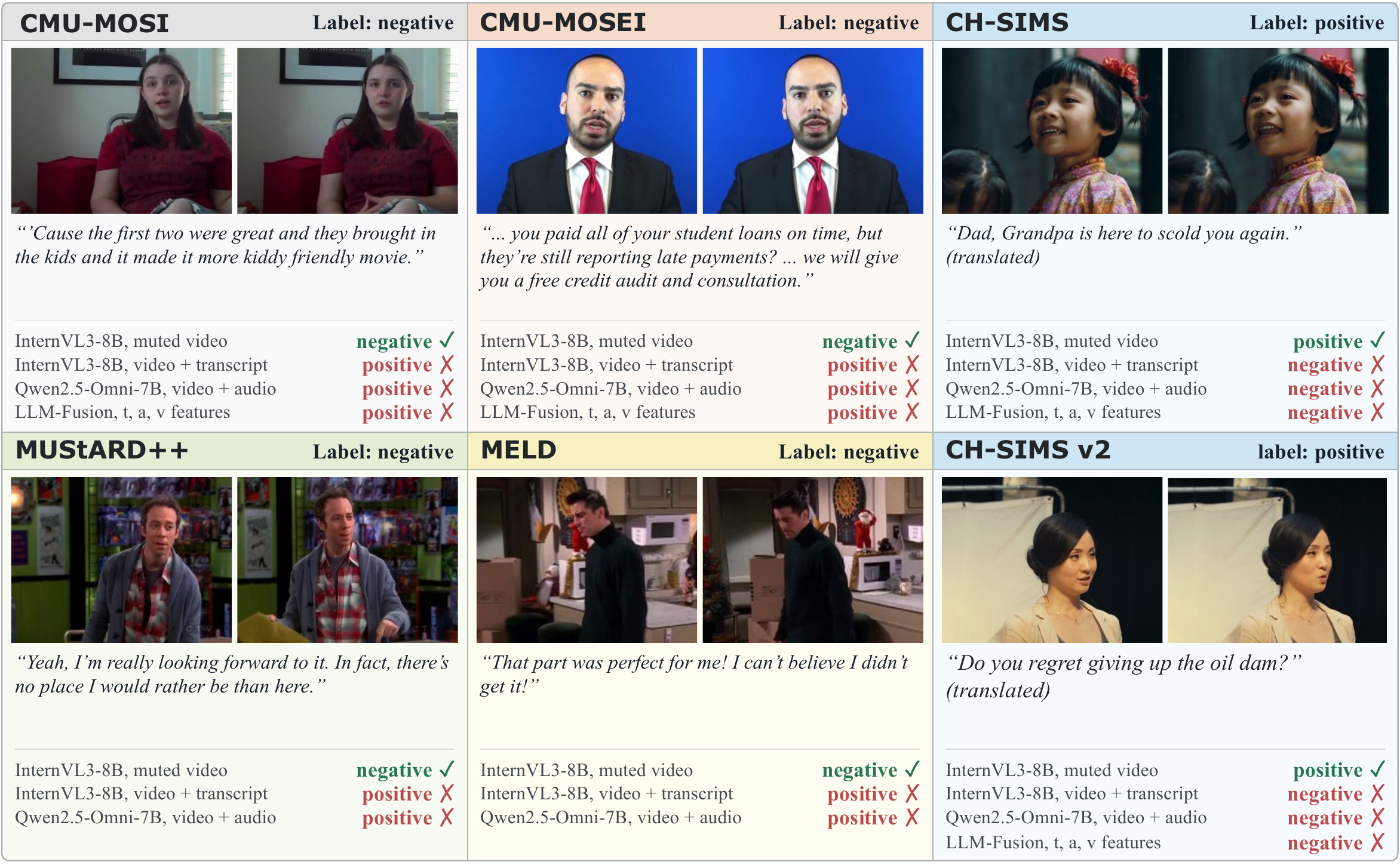}\par
\vspace{2pt}
\captionsetup{hypcap=false}
\captionof{figure}{\textbf{Language-shortcut clips from MR-Bench-H on which current models follow the words.} Every
model is asked whether the speaker's sentiment is positive or negative: a vision--language model that answers correctly
from the muted video (green \cmark) switches to the polarity of the words once the transcript is added (red \xmark), and
an audio-visual MLLM and the fusion backbone LLM-Fusion (trained on the four MMSA sources only) follow the words as
well.}
\captionsetup{hypcap=true}
\label{fig:teaser}
\end{strip}
\begin{abstract}\enlargethispage{-\baselineskip}
Is a face or a voice still heard once the words are known? Incomplete multimodal learning increasingly uses large
language models (LLMs) as fusion engines: modality features enter as token blocks and missing modalities become
placeholders. We find that LLM-driven fusion can override non-verbal evidence it can read. Contrasting its full-input
prediction with \emph{counterfactual views}, the same model shown one modality or no content, we identify language
shortcuts. We build MR-Bench: MR-Bench-A types every test sample of four sentiment benchmarks; MR-Bench-H contains
900 human-validated clips (620 language-shortcut, LS; 280 non-verbal-shortcut, NS) from seven sources, and every
model is scored on all of them. Zero-shot MLLMs solve at most 32.7\% of the LS clips;
adding the transcript to the muted video lowers vision--language accuracy by 46.1--53.1 points. Yet inverting a text-only
model's answers solves 94.4\%, so LS accuracy alone cannot establish use of non-verbal evidence. Inside the
fusion LLM, fused states align increasingly with the text view late in the decoder. Counterfactual-View Reliability learning (CoVeR) trains views as unimodal judges and
withholds fused supervision where they disagree. It achieves state-of-the-art results on five benchmarks without
extra inference cost in accuracy mode. On MR-Bench-H, it reaches 61.7\% LS and 93.6\% NS accuracy; robust mode raises
LS accuracy to 63.0\% while lowering NS accuracy to 84.8\%. Among LS clips solved without text, adding text causes
errors on 30.1\% with CoVeR versus 32.2\% with LLM-Fusion.
Project page: \url{https://peacelwh.github.io/CoVeR/}.
\end{abstract}
\section{Introduction}
\label{sec:intro}

Multimodal signals in the wild are rarely complete, and the causes that corrupt test data corrupt training data as
well. Most incomplete multimodal learning (IML) methods reconstruct the missing modalities~\cite{mmin,tfrnet,gcnet,imder,dicmor}
or distil from a complete-modality teacher~\cite{corrkd,momke}, and therefore still need complete training observations.
Large language models (LLMs) offer an alternative: modality features enter as token blocks, missing modalities become
learnable placeholders, and a prompt describes availability. We study such \emph{LLM-driven fusion} under dynamic
modality masking, with a pattern sampled at each training visit.

Yet these models still fail on cases that are elementary for a human. In the clips of \cref{fig:teaser}, a speaker
sarcastically says ``there's no place I would rather be than here'', and a child announces with a beaming smile that
grandpa has come to scold their father. A vision--language model that answers correctly from the muted video switches to
the polarity of the words once the transcript is added; an audio-visual MLLM and an LLM-driven fusion model follow the
words as well. We ask: \emph{when the observed modalities disagree, whose evidence does the model follow, and where does
this failure originate?} Our hypothesis is that the modalities do not enter LLM-driven fusion on equal footing: the
decoder is pretrained on text, text carries most of the training signal of sentiment corpora~\cite{almt,lnln}, and
audio and vision arrive only as projected features, so the text prior may turn modality
imbalance~\cite{wang2020makes,huang2022modality} into a shortcut that overrides the other modalities whenever text is
present, as language priors do in visual question answering~\cite{vqacp,rubi}.

\begin{figure*}[t]
\centering
\includegraphics[width=0.85\textwidth]{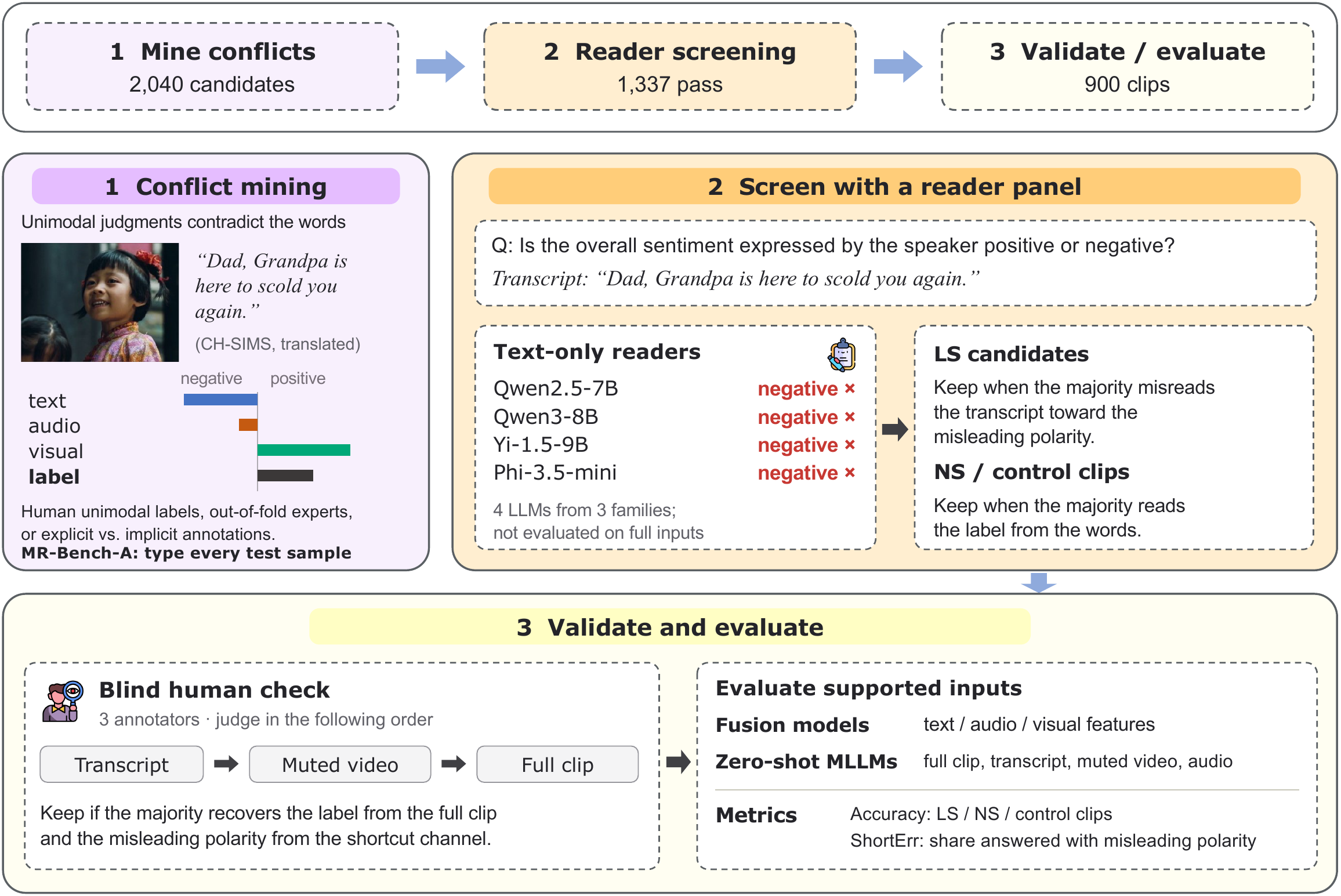}
\vspace{-4pt}
\caption{\textbf{Constructing MR-Bench.} Candidates are clips whose unimodal judgments disagree with the words (here a
CH-SIMS clip with human unimodal labels on $[-1,1]$); a panel of four text-only LLMs keeps a language-shortcut (LS)
candidate only if the majority misreads its transcript, and three annotators keep a clip only if the majority recovers
the label from the full clip and the misleading polarity from the shortcut channel. Models are then asked the same
binary question from every input they support.}
\label{fig:pipeline}
\vspace{-8pt}
\end{figure*}

To test this hypothesis, we exploit a property of LLM-driven fusion: because every modality occupies a token block at a
fixed position, the \emph{same} model can be shown a single modality, or no content at all, by swapping blocks for
placeholders; these \emph{counterfactual views} are aligned with the full input token by token and cost one batched
forward pass. We call inputs on which the fused prediction follows the words while the non-verbal evidence and the label
say otherwise \emph{language-shortcut cases}, and find them without relying on the model under study, through unimodal
judgments: human labels of what each modality conveys on its own~\cite{chsims,chsimsv2} or out-of-fold single-modality
experts. From these cases we build MR-Bench (\cref{fig:pipeline}): MR-Bench-A types every test sample of
CMU-MOSI~\cite{mosi}, CMU-MOSEI~\cite{mosei}, CH-SIMS~\cite{chsims} and CH-SIMSv2~\cite{chsimsv2} by its unimodal
judgments, and MR-Bench-H contains 900 human-validated clips (620 LS, 280 NS) from seven sources, including
sarcasm~\cite{mustardpp}, multi-party dialogue~\cite{meld} and acted dyadic interaction~\cite{iemocap}. Clips are mined
with unimodal judgments, screened by text-only LLMs and validated by three annotators. Every item asks whether the
speaker's sentiment is positive or negative. Every model is scored on all 900 clips: the strongest zero-shot MLLM
solves 32.7\% of the LS clips (\cref{fig:results}). Adding the transcript
to the muted video lowers vision--language accuracy by 46.1--53.1 points. Since text-polarity inversion alone solves
94.4\% of the LS clips, LS scores must be interpreted with NS accuracy and paired views.

We then analyse the LLM-fusion backbone itself (\cref{sec:findings}). With text present, it is right on 90--98\% of the
cases on which all modalities agree but only on 17--59\% of the conflict cases of MR-Bench-A, although the same model
solves 63--87\% of them once the text is removed. Late-layer fused states align increasingly with the text view
(prior-centred cosine 0.68--0.81), and a simple cosine-discrepancy gate separates conflict from consensus cases
hardly better than chance. Scaling from 0.5B to 1.7B parameters does not resolve
the failure. The views' prediction-level disagreement provides a useful signal. We study
two ways of using this signal (\cref{fig:overview}): letting disagreeing views override the fused decision at test time
trades consensus accuracy and correlation for conflict accuracy, because untrained views are weak judges, whereas
Counterfactual-View Reliability learning (CoVeR) trains the views as unimodal judges and withholds fused supervision
wherever they disagree. CoVeR keeps the architecture and the inference cost of the backbone, improves it on five
benchmarks, achieves state-of-the-art results on all five against nine IML methods re-run under one protocol, and
is evaluated for shortcut-following error on MR-Bench-A. On MR-Bench-H, CoVeR in accuracy mode reaches 61.7\% LS and 93.6\%
NS accuracy. On paired LS views, the fraction of AV-correct clips made incorrect by adding text drops from
32.2\% to 30.1\%. Our contribution is objective-level conflict handling for feature-input LLM fusion: training
unimodal judges and suppressing incompatible fused supervision, complementing inference-time interventions.
\section{The MR-Bench Benchmark}
\label{sec:bench}

\subsection{Finding modality-conflict cases}
\label{sec:conflict}
\label{sec:prelim}

\para{LLM-driven fusion.}
A sample consists of modality features $X=\{X^{j}\}_{j\in\M}$, $\M=\{t,a,v\}$ (text, audio, visual),
$X^{j}\in\mathbb{R}^{T\times d_j}$, an availability mask $\mathbf{m}\in\{0,1\}^{3}\setminus\{\mathbf 0\}$ and a
sentiment score $y$. Observations are incomplete \emph{at training time}: a missing modality is unavailable both as input and as a target.
The main protocol dynamically samples one of the seven non-empty patterns at each visit, including the complete
pattern; it does not make a sample's missingness permanent. Permanent missingness, where every training and
validation sample keeps one pattern and its unobserved modalities are never read, is evaluated separately
(\cref{sec:s-fixed}). LLM-Fusion serialises a sample into
\begin{equation}
Z(\mathbf c,\mathbf p)=\big[P(\mathbf p);\,S_{t}(c_t);\,S_{a}(c_a);\,S_{v}(c_v);\,E_{\mathrm{fus}}\big],
\label{eq:seq}
\end{equation}
where $P(\mathbf p)$ is a prompt stating which modalities are available, $S_j(1)=[\langle j_s\rangle,\phi_j(X^{j}),\langle j_e\rangle]$
is a projected feature block with boundary tokens, $S_j(0)$ replaces $\phi_j(X^{j})$ by $T$ copies of a learnable
$\langle\textit{miss}_j\rangle$ embedding, and $E_{\mathrm{fus}}$ are $K$ fusion tokens. A LoRA-tuned decoder-only LLM
$f_\theta$ produces fusion-token states $H_\ell\in\mathbb R^{K\times D}$ at every layer $\ell$, and a mask-aware
dual-path head decodes the last-layer states into the score $\hat y$. The model is trained with
$\mathcal L_{\mathrm{L}}=\lambda_{\mathrm{task}}\,\ell(\hat y,y)+\lambda_{\mathrm{con}}\mathcal L_{\mathrm{con}}
+\lambda_{\mathrm{reg}}\mathcal L_{\mathrm{reg}}$, where $\ell$ is the squared error, $\mathcal L_{\mathrm{con}}$ ties the
two decoding paths and $\mathcal L_{\mathrm{reg}}$ keeps the fusion tokens diverse; for an observed sample,
$\mathbf c=\mathbf p=\mathbf m$.

\para{Counterfactual views.}
Decoupling the two masks yields counterfactual inputs to the \emph{same} model: a \emph{view} $\V=(\mathbf c,\mathbf p)$
specifies which blocks carry features and what the prompt claims. For a sample with availability $\mathbf m$ we use the
full view $\V_{f}=(\mathbf m,\mathbf m)$, the single-modality views $\V_{j}=(\mathbf e_j\odot\mathbf m,\
\mathbf e_j\odot\mathbf m)$ for every observed $j$, and the content-free view $\V_{\varnothing}=(\mathbf 0,\mathbf m)$,
which keeps the prompt but hides all content. All views share the token layout of \cref{eq:seq}, so their fusion-token states $H^{\V}_{\ell}$ are aligned position by
position and run in one batched pass; $\hat y$, $\hat y^{(j)}$ and $\hat y^{(\varnothing)}$ denote their predictions.

\para{Unimodal judgments and conflict cases.}
A view tells us what the model makes of one modality; what the modality actually conveys must come from a judgment
independent of the model under study. Every sample therefore receives a judgment $u^{(j)}$ per modality: the human
unimodal labels on CH-SIMS/CH-SIMSv2~\cite{chsims,chsimsv2} and, on CMU-MOSI/MOSEI, the predictions of single-modality
BiLSTM experts (three seeds, out-of-fold on the training split). A judgment is \emph{confident} if $|u^{(j)}|\ge\kappa$
($\kappa=0.2$ on the $[-1,1]$ scale of CH-SIMS, $0.5$ on the $[-3,3]$ scale of CMU-MOSI/MOSEI). Non-neutral samples are
typed as \emph{consensus} (all judgments share the polarity of the label), \emph{conflict} (the text judgment is
confidently wrong and an audio or visual judgment confidently right), \emph{weak-modality conflict} (text right, audio
or visual confidently wrong) or \emph{other}. A conflict case on which a model predicts the polarity of the text
judgment is a \emph{language-shortcut} error; ShortErr is the fraction of conflict cases on which a model makes it.

\subsection{Designing MR-Bench from conflict cases}
\label{sec:design}

\para{MR-Bench-A: every test sample, typed.}
MR-Bench-A applies this typing to the complete test splits of CMU-MOSI, CMU-MOSEI, CH-SIMS and CH-SIMSv2 (counts in
\cref{tab:bench}, supplementary material), so that any model trained on them can report Acc2 per case type and ShortErr
next to its usual metrics; on CH-SIMS/CH-SIMSv2 the conflict sets are defined by human annotation alone.

\para{MR-Bench-H: language-shortcut clips.}
MR-Bench-A is exhaustive, but its conflict sets are small and model-judged on CMU-MOSI/MOSEI. MR-Bench-H therefore
collects raw clips from seven sources in which a clear label meets a \emph{shortcut channel} that suggests the opposite
polarity: the words for \emph{language-shortcut} (LS) clips (sarcasm, polite refusal, ironic report), a salient facial
expression or gesture for \emph{non-verbal-shortcut} (NS) clips, and no shortcut for control clips. It is built in three
steps (\cref{fig:pipeline}): (i)~\emph{mining} candidates from the unimodal judgments of MR-Bench-A, the explicit \vs
implicit emotion annotations of MUStARD++~\cite{mustardpp} and, for MELD~\cite{meld} and IEMOCAP~\cite{iemocap}, the
text-only view of a trained fusion model; (ii)~\emph{screening} by a panel of four text-only LLMs from three families,
which keeps an LS candidate only if the majority misreads its transcript and an
NS candidate or control only if the majority reads the label from the words; (iii)~\emph{blind human validation} by
three annotators, who judge each clip from the transcript, the muted video and the full clip, in this order. Of 2,040
candidates from validation and test splits and unseen corpora, 1,337 pass the panel and 900 the annotators (620 LS, 280
NS; Fleiss' $\kappa=0.75$). Each clip asks one deliberately simple question, ``Is the overall sentiment expressed by
the speaker positive or negative?'', so that a failure cannot be attributed to the difficulty of the task (per-source
statistics in the supplementary material).

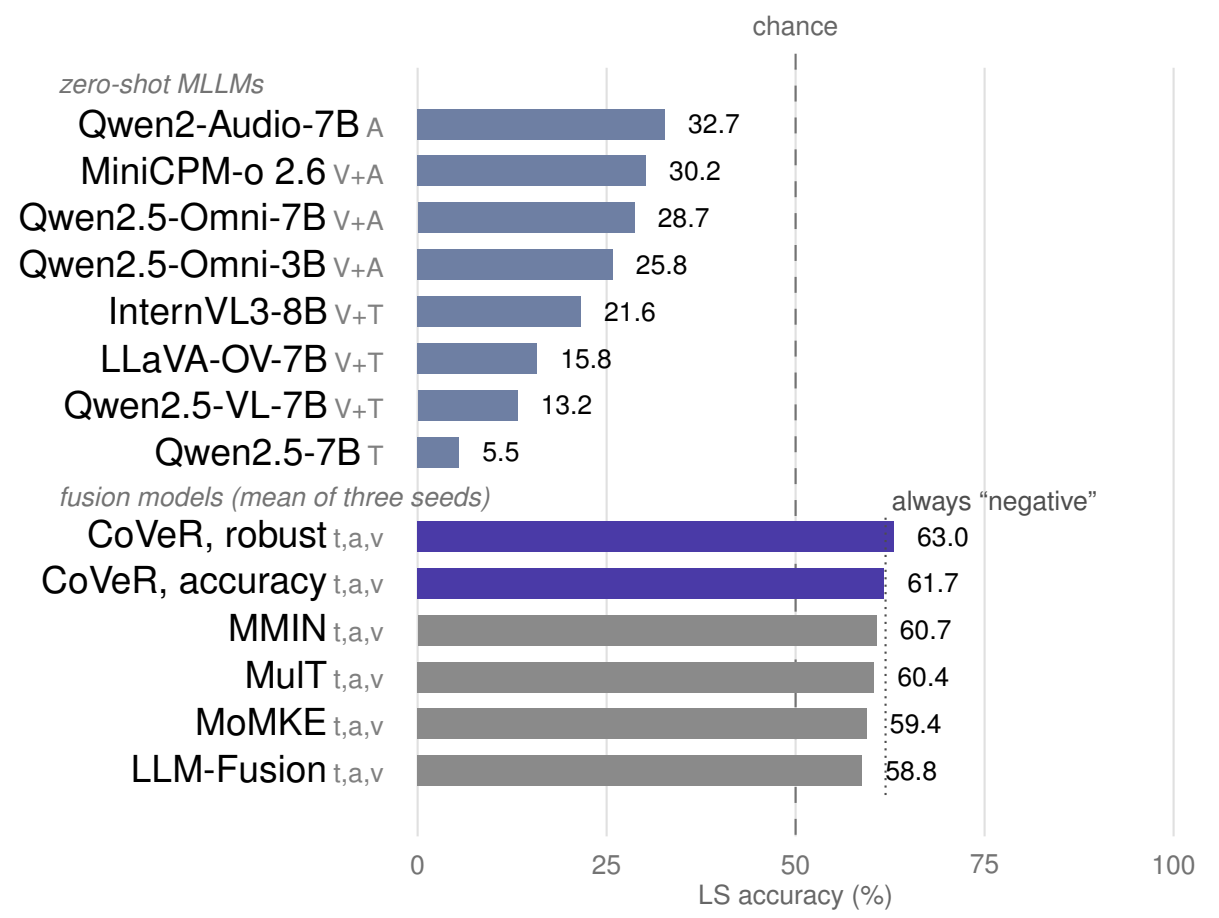
\begin{figure}[t]
\centering
\begin{tikzpicture}[font=\sffamily\scriptsize,x=0.05cm,y=0.31cm]
\def\lab{-2}
\foreach \t in {0,25,50,75,100}{
\draw[black!12] (\t,0) -- (\t,-16.4);
\node[black!55,font=\sffamily\tiny] at (\t,-17.0) {\t};}
\draw[black!55,dashed] (50,0.3) -- (50,-16.4);
\node[black!60,font=\sffamily\tiny,anchor=south] at (50,0.3) {chance};
\node[black!60,font=\sffamily\tiny] at (50,-17.7) {LS accuracy (\%)};
\newcommand{\resbar}[5]{
\node[anchor=east] at (\lab,-#1) {#2\,{\color{black!50}\tiny#3}};
\fill[#5] (0,-#1-0.33) rectangle (#4,-#1+0.33);
\node[anchor=west,font=\sffamily\tiny] at (#4+0.6,-#1) {#4};}
\node[anchor=west,black!55,font=\sffamily\tiny\itshape] at (\lab-48,-0.35) {zero-shot MLLMs};
\resbar{1.2}{Qwen2-Audio-7B}{A}{32.7}{cmllm}
\resbar{2.2}{MiniCPM-o 2.6}{V+A}{30.2}{cmllm}
\resbar{3.2}{Qwen2.5-Omni-7B}{V+A}{28.7}{cmllm}
\resbar{4.2}{Qwen2.5-Omni-3B}{V+A}{25.8}{cmllm}
\resbar{5.2}{InternVL3-8B}{V+T}{21.6}{cmllm}
\resbar{6.2}{LLaVA-OV-7B}{V+T}{15.8}{cmllm}
\resbar{7.2}{Qwen2.5-VL-7B}{V+T}{13.2}{cmllm}
\resbar{8.2}{Qwen2.5-7B}{T}{5.5}{cmllm}
\node[anchor=west,black!55,font=\sffamily\tiny\itshape] at (\lab-48,-9.2) {fusion models (mean of three seeds)};
\resbar{10.0}{CoVeR, robust}{t,a,v}{63.0}{cours}
\resbar{11.0}{CoVeR, accuracy}{t,a,v}{61.7}{cours}
\resbar{12.0}{MMIN}{t,a,v}{60.7}{cbase}
\resbar{13.0}{MulT}{t,a,v}{60.4}{cbase}
\resbar{14.0}{MoMKE}{t,a,v}{59.4}{cbase}
\resbar{15.0}{LLM-Fusion}{t,a,v}{58.8}{cbase}
\draw[black!70,densely dotted] (61.9,-9.6) -- (61.9,-15.6);
\node[black!70,font=\sffamily\tiny,anchor=south west,inner sep=1pt] at (61.9,-9.7) {always ``negative''};
\end{tikzpicture}
\vspace{-4pt}
\caption{\textbf{MR-Bench-H results} on all 620 LS clips of the released set (seven sources; \cref{tab:s-mrbh}). T: transcript, A: audio, V: video. Fusion models are trained on the training split of each source and averaged over three seeds; they inherit its label prior (dotted: the constant answer ``negative'', 61.9\%), so they are compared within their family. Every zero-shot MLLM is far below chance; inverting a text-only model's answers alone reaches 94.4\%, so LS accuracy is read together with the 280 NS clips.}
\label{fig:results}
\vspace{-8pt}
\end{figure}
\subsection{Benchmark results}
\label{sec:survey}

We evaluate zero-shot MLLMs that read text (Qwen2.5-7B~\cite{qwen25}), hear audio
(Qwen2-Audio-7B~\cite{qwen2audio}), watch video with transcripts (LLaVA-OneVision-7B~\cite{llavaov},
Qwen2.5-VL-7B~\cite{qwen25vl}, InternVL3-8B~\cite{internvl3}) or watch and hear clips
(MiniCPM-o~2.6~\cite{minicpmv}, Qwen2.5-Omni-3B/7B~\cite{qwen25omni}), alongside trained fusion models.
Every model is scored on all 900 clips of MR-Bench-H (620 LS, 280 NS). MLLMs take the raw clips; a fusion model
reads features, so each clip is scored by the model trained on the training split of its source corpus, with
features extracted by us for IEMOCAP, MELD and MUStARD++ (three seeds; \cref{sec:s-mrbh}).

\para{Models follow the words.}
Zero-shot MLLMs solve 5.5--32.7\% of the LS clips (\cref{fig:results}), all far below chance, although they solve
89.6--99.3\% of the NS clips, whose words carry the label. ShortErr, the share of LS clips answered with the
misleading polarity of the words, is 66.9--94.4\%. Fusion models are trained on the source corpora and inherit
their label priors and in-domain words (the constant answer ``negative'' solves 61.9\% of the LS clips), so they are
compared within their family: CoVeR reaches 61.7\% LS and 93.6\% NS accuracy in accuracy mode, versus
58.8\% and 93.1\% for LLM-Fusion, and robust mode 63.0\% LS but 84.8\% NS (\cref{tab:s-mrbh}).

\para{LS accuracy alone is insufficient.}
The text-only model solves 34 of the 620 LS clips (5.5\%); reversing its answers solves 94.4\% of them without
audio or video, but only 0.7\% of the NS clips. LS accuracy alone therefore cannot prove the use of non-verbal
evidence: we read it together with NS accuracy and with paired views of the same model
(\cref{fig:words,tab:s-mrbh}).
\section{Systematic Failures of LLM-Driven Fusion}
\label{sec:findings}
\label{sec:diag}

Models fail on MR-Bench, and the clips suggest why: they follow the words. To find where this failure originates, we
turn to the backbone itself, LLM-Fusion trained with $\mathcal L_{\mathrm L}$ (\cref{sec:setup}); its MR-Bench-A
statistics are listed in \cref{tab:bench} (supplementary material).

\subsection{Conflict patterns in MR-Bench-A}
\label{sec:patterns}

\para{The model fails exactly where the modalities disagree, although the evidence is there.}
With all modalities observed, LLM-Fusion is right on 90--98\% of the consensus cases but only on 17--59\% of the conflict
cases, and 41--83\% of its predictions on these cases follow the polarity of the text (\cref{tab:bench}). The failure
is not a lack of evidence: the \emph{same} model, evaluated without text, solves 63--87\% of the conflict cases, and its
own visual-only view solves 55--84\%. The severity grows as the non-dominant features weaken.

\begin{figure*}[t]
\centering
\includegraphics[width=\textwidth]{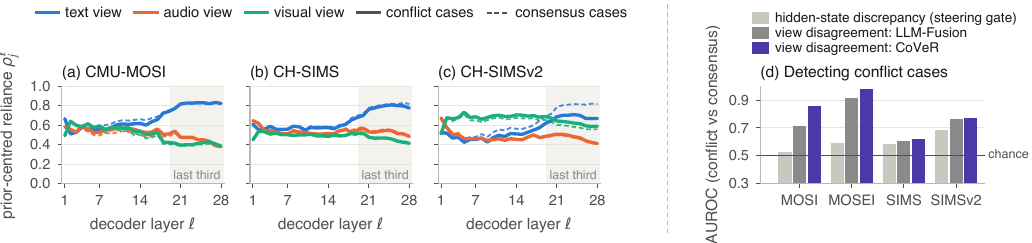}
\vspace{-10pt}
\caption{\textbf{Where and how the decision collapses.} (a--c) Prior-centred reliance $\rho^{\ell}_{j}$
(\cref{eq:reliance}) of the fused state on the text-, audio- and visual-only views across the decoder layers of
LLM-Fusion, on conflict (solid) and consensus (dashed) cases. (d) AUROC for separating conflict from consensus cases
with the hidden-state discrepancy $1-\cos(h^{f},h^{t})$ that gates activation steering and with the prediction-level
disagreement between the audio/visual and the text views. Mean over 3 seeds.}
\label{fig:diag}
\vspace{-6pt}
\end{figure*}

\subsection{Where the decision collapses}
\label{sec:collapse}

Since the prompt and the special tokens dominate the states, raw cosine similarities between views are close to one
at every layer; we therefore measure how much the content-driven part of the fused state aligns with that of a
single-modality view,
\begin{equation}
\rho^{\ell}_{j}=\cos\!\big(\bar h^{f}_{\ell}-\bar h^{\varnothing}_{\ell},\ \bar h^{j}_{\ell}-\bar h^{\varnothing}_{\ell}\big),
\label{eq:reliance}
\end{equation}
where $\bar h^{\V}_{\ell}$ is the mean of $H^{\V}_{\ell}$ over the fusion tokens; $\rho^{\ell}_{j}\to1$ means that the
content-driven components become directionally aligned; this is not a measure of causal contribution.

\para{Collapse forms late and is hierarchical.}
Text reliance averages 0.49--0.63 over the first two thirds of the decoder and rises to 0.68--0.81 in the last third,
while visual reliance falls (\cref{fig:diag}a--c). Without text, the fused state aligns with the audio-only view instead
($\rho_a=0.93$ \vs $\rho_v=0.38$ on CMU-MOSI in the last third), \ie the modalities are ranked $t\succ a\succ v$. We call
this tendency \emph{dominant-modality collapse}: increasing alignment with the dominant view near readout,
not equality of the fused and text-only states.

\para{State discrepancy is a weak conflict signal.}
Text reliance on conflict cases is as high as on consensus cases; only CH-SIMSv2, which has the most training
conflicts, partly registers the disagreement (\cref{fig:diag}c). Accordingly, the discrepancy $d=1-\cos(h^{f},h^{t})$,
which can gate activation steering~\cite{iti,caa} of the full view away from a shortcut view, separates conflict from
consensus cases hardly better than chance (AUROC 0.53--0.60 on CMU-MOSI, CMU-MOSEI and CH-SIMS; \cref{fig:diag}d), and
its gate ($d>0.15$) fires on at most 4\% of the text-present test samples and leaves ShortErr unchanged
(\cref{tab:reliability}).

\para{Prediction disagreement supplies a useful signal.}
In contrast, the \emph{predictions} of the views are informative: the visual-only view disagrees in polarity with the
text-only view on 73\% of the conflict cases but only on 27\% of the consensus cases (CH-SIMSv2), and this disagreement
separates the two case types with an AUROC of up to 0.92 (\cref{fig:diag}d). The non-dominant views, however, are weak
judges, because nothing trains them to be faithful to their modality.

\subsection{Scaling the LLM, and zero-shot MLLMs}
\label{sec:scaling}
\label{sec:mllm}

A larger language model does not solve the problem: with five decoder-only LLMs from 0.5B to 1.7B parameters on
CH-SIMS, everything else fixed (\cref{tab:scaling}), the correlation of LLM-Fusion stays at
48.9--49.9, its conflict accuracy varies between 35.5\% and 46.8\% without a consistent trend in size, and every
backbone still follows the text on 53--65\% of the conflict cases although it solves 63--73\% of them once the text is
removed. On the 620 LS clips of MR-Bench-H, Qwen2.5-Omni-7B solves 178 versus 160 for its 3B sibling (28.7\% \vs
25.8\%). Paired views of Qwen2.5-VL, InternVL3 and LLaVA-OV show drops of 47.9, 46.1 and 53.1 points when the
transcript is added to the muted video (\cref{fig:words}, supplementary material). This supports a
transcript-induced failure on these clips, while the text-inversion control cautions against interpreting LS accuracy
alone as evidence of multimodal reasoning. Scaling alone leaves substantial shortcut errors.
\section{Counterfactual-View Reliability Learning}
\label{sec:method}
\begin{figure*}[t]
\centering
\includegraphics[width=0.85\textwidth]{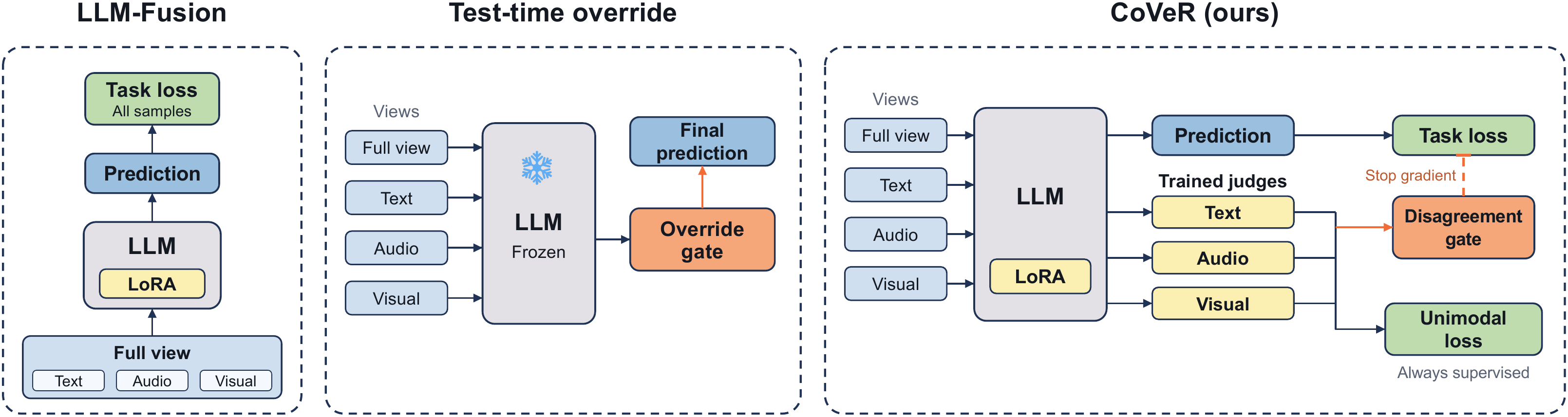}
\vspace{-6pt}
\caption{\textbf{Three ways to fuse.} LLM-Fusion supervises the fused prediction on every sample (left); the test-time
override lets confidently disagreeing views of the frozen model replace the fused decision (middle, \cref{eq:gate,eq:mix});
CoVeR trains the views as unimodal judges and withholds the fused task loss wherever a judge disagrees with the text
judge (right), running the full view alone at test time.}
\label{fig:overview}
\vspace{-8pt}
\end{figure*}

The prediction-level disagreement of counterfactual views provides a direct signal of conflicting evidence
(\cref{sec:findings}). We use this signal in two ways (\cref{fig:overview}): letting disagreeing views override the fused decision at
test time, without any training (\cref{sec:override}), and Counterfactual-View Reliability learning (CoVeR), which turns
the views into trained judges whose disagreement decides which samples supervise the fused decision (\cref{sec:cover}).

\subsection{Experiment setting}
\label{sec:setup}
We use CMU-MOSI~\cite{mosi}, CMU-MOSEI~\cite{mosei}, CH-SIMS~\cite{chsims} and CH-SIMSv2~\cite{chsimsv2} with the
standard MMSA features~\cite{msena}, pooled to $T=50$ tokens per modality, and IEMOCAP~\cite{iemocap}
(Session 5 test), scored by its averaged valence, with BERT, openSMILE ComParE and ViT facial-expression features
extracted by us and its 512 MR-Bench-H candidates held out. Training observations are incomplete (\cref{sec:conflict}); at test time we
evaluate every availability pattern and random missing rates $r\in\{0.1,\dots,0.7\}$ and report Acc2, weighted F1, MAE
and Pearson correlation (Corr), averaged over the patterns (Avg) or the missing rates (Rate), and the MR-Bench metrics
with all modalities observed. Nine IML methods are ported from the authors' code and run under our protocol
(same frozen features, availability patterns and tests); methods whose objectives use the features of missing
modalities as reconstruction or distillation targets keep this access (T-Miss \xmark). LLM-Fusion is the backbone of \cref{sec:conflict} trained
with $\mathcal L_{\mathrm L}$ alone, \ie CoVeR without judges and gate; both use Qwen3-0.6B~\cite{qwen3} with
LoRA~\cite{lora}, $K=4$ fusion tokens, $\lambda_{\mathrm{task}}=10$, $\lambda_{\mathrm{con}}=\lambda_{\mathrm{reg}}=1$,
AdamW~\cite{adamw}, batch 32 and a default budget of 30 epochs (15 on CMU-MOSEI) on one 24\,GB GPU. Hyper-parameters ($\tau$,
$\lambda_{\mathrm{uni}}$, LoRA rank, learning rate, epochs, weight averaging, regression loss) are selected per dataset
on the validation split only; the analyses of \cref{tab:reliability,tab:ablation} use the default recipe of the
backbone for every row ($\tau=0.3$, $\lambda_{\mathrm{uni}}=1$). Means over three seeds for the comparisons with the
state of the art and the diagnosis of \cref{sec:findings}, over five seeds for the analyses of CoVeR; search ranges,
selected values, baseline deviations and feature extraction are in the supplementary material.

\subsection{Test-time override}
\label{sec:override}
The most direct use of the views is to let them overrule the fused decision. Let the \emph{anchor} be the dominant
modality (text on all four benchmarks) and $c$ the neutral score. A non-anchor view $j\in\M\setminus\{t\}$
\emph{confidently disagrees} with the anchor if
\begin{equation}
\begin{aligned}
g_j={}&\1\big[\sgn(\hat y^{(j)}-c)\neq\sgn(\hat y^{(t)}-c)\big]\\
&\cdot\1\big[\,|\hat y^{(j)}-c|\ge\tau/2\,\big],
\end{aligned}
\label{eq:gate}
\end{equation}
with a margin $\tau$ ($\hat y^{(t)}$ is the content-free prediction when $m_t=0$). With budgeted weights
$w_j=m_j g_j/\max(1,\sum_k m_k g_k)$, the gated prediction is
\begin{equation}
\tilde y=\hat y+\sum\nolimits_{j} w_j\big(\sg[\hat y^{(j)}]-\hat y\big),
\label{eq:mix}
\end{equation}
where $\sg[\cdot]$ stops the gradient: if no view disagrees, $\tilde y=\hat y$; otherwise $\tilde y$ is the mean of the
disagreeing views' predictions. Applied to a trained LLM-Fusion with a test-time margin $\tau_{\mathrm{test}}$,
\cref{eq:mix} needs no training and one extra batched pass over the views.

\para{Overriding trades consensus for conflict accuracy.}
A smaller margin raises conflict accuracy but lowers consensus accuracy and correlation. On CMU-MOSI,
\cref{tab:reliability} shows conflict accuracy rising from 21.7\% to 33.9\%, while consensus accuracy falls from
94.1\% to 92.5\% and Avg Corr from 50.7 to 47.8. The separate margin sweeps in \cref{tab:override} show the
same trade-off. Untrained views remain imperfect judges: their disagreement can reflect an error in either view.

\begin{table*}[t]
\centering
\caption{\textbf{Correcting the collapse on MR-Bench-A.} Avg Corr ($\times100$), Acc2 on consensus (Cons.) and
conflict (Conf.) cases and ShortErr (SE), in \%. T-free: no training; +Views: extra test-time pass over the views.
Mean over 5 seeds; details in the supplementary material.}
\label{tab:reliability}
\footnotesize
\setlength{\tabcolsep}{5.5pt}
\begin{tabular}{@{}lcc cccc cccc@{}}
\toprule
\multirow[c]{2}{*}[-2.4pt]{Method} &\multirow[c]{2}{*}[-2.4pt]{T-free} &\multirow[c]{2}{*}[-2.4pt]{+Views} & \multicolumn{4}{c}{CMU-MOSI} & \multicolumn{4}{c}{CH-SIMSv2}\\
\cmidrule(lr){4-7}\cmidrule(l){8-11}
 & & & Corr\up & Cons.\up & Conf.\up & SE\dn & Corr\up & Cons.\up & Conf.\up & SE\dn\\
\midrule
LLM-Fusion                            & --     & -- & 50.7&94.1&21.7&78.3&59.2&93.7&58.6&41.4\\
\quad + discrepancy-gated steering & \cmark & \cmark & 50.3&93.8&22.2&77.8&58.9&93.6&58.6&41.4\\
\quad + test-time override ($\tau_{\mathrm{test}}{=}0.3$) & \cmark & \cmark & 47.8&92.5&33.9&66.1&55.3&86.8&72.3&27.7\\
\quad + learned reliability gate            & \xmark & \cmark & 50.1&92.5&19.4&80.6&58.6&93.1&56.4&43.6\\
\quad + conflict re-weighting                & \xmark & \xmark & 50.4&93.6&24.4&75.6&58.1&93.0&52.3&47.7\\
\quad + unimodal multi-task ($\Luni$)        & \xmark & \xmark & 52.2&95.7&24.4&75.6&61.3&93.9&55.6&44.4\\
\midrule
\rowcolor{oursrow}
CoVeR, accuracy mode                         & \xmark & \xmark & 53.5&95.2&28.3&71.7&61.2&93.7&59.5&40.5\\
\rowcolor{oursrow}
CoVeR, robust mode ($\tau_{\mathrm{test}}{=}0.3$) & \xmark & \cmark & 51.9&95.3&31.1&68.9&57.4&87.6&69.9&30.1\\
\bottomrule
\end{tabular}
\vspace{-6pt}
\end{table*}

\begin{table*}[t]
\centering
\caption{\textbf{Comparison with the state of the art under incomplete modalities}, averaged over the seven test-time
availability patterns (Acc2 and F1 in \%). T-Miss \xmark: the features of missing
modalities serve as reconstruction or distillation targets. Best / second / third highlighted;
mean over 3 seeds. Deltas use displayed means and are approximate.}
\label{tab:sota}
\scriptsize
\setlength{\tabcolsep}{2.4pt}
\begin{tabular}{@{}llc cccc cccc cccc cccc@{}}
\toprule
\multirow[c]{2}{*}[-2.4pt]{Method} &\multirow[c]{2}{*}[-2.4pt]{Venue} &\multirow[c]{2}{*}[-2.4pt]{T-Miss} & \multicolumn{4}{c}{CMU-MOSI} & \multicolumn{4}{c}{CMU-MOSEI} & \multicolumn{4}{c}{CH-SIMS} & \multicolumn{4}{c}{CH-SIMSv2}\\
\cmidrule(lr){4-7}\cmidrule(lr){8-11}\cmidrule(lr){12-15}\cmidrule(l){16-19}
 & & & Acc2\up & F1\up & MAE\dn & Corr\up & Acc2\up & F1\up & MAE\dn & Corr\up & Acc2\up & F1\up & MAE\dn & Corr\up & Acc2\up & F1\up & MAE\dn & Corr\up\\
\midrule
MulT~\cite{mult} & ACL'19 & \cmark
& \cellcolor{second}68.6 & 65.5 & 1.160 & \cellcolor{second}0.521
& 73.9 & 68.9 & 0.706 & 0.532
& 72.6 & 69.6 & 0.510 & 0.414
& 73.8 & 71.5 & 0.393 & 0.534 \\
MISA~\cite{misa} & MM'20 & \cmark
& 67.0 & 65.4 & 1.166 & 0.479
& 74.4 & 72.8 & 0.702 & 0.549
& 72.6 & 68.7 & 0.520 & 0.463
& 73.6 & 72.7 & 0.397 & 0.550 \\
MMIN~\cite{mmin} & ACL'21 & \xmark
& 67.7 & \cellcolor{third}67.0 & 1.205 & 0.504
& \cellcolor{second}76.3 & 74.8 & 0.690 & \cellcolor{third}0.574
& 74.6 & 74.4 & 0.502 & 0.508
& 75.2 & 74.3 & 0.388 & 0.577 \\
TFR-Net~\cite{tfrnet} & MM'21 & \xmark
& 67.6 & 63.8 & \cellcolor{third}1.140 & 0.501
& 73.1 & 67.0 & 0.696 & 0.515
& 73.7 & 71.3 & 0.508 & 0.462
& 74.6 & 73.5 & 0.382 & 0.565 \\
IMDer~\cite{imder} & NeurIPS'23 & \xmark
& 68.3 & 65.9 & 1.151 & 0.506
& 75.2 & 70.7 & \cellcolor{second}0.675 & 0.567
& 75.3 & 72.4 & 0.495 & 0.478
& \cellcolor{third}75.7 & 75.3 & \cellcolor{third}0.371 & \cellcolor{third}0.581 \\
DiCMoR~\cite{dicmor} & ICCV'23 & \xmark
& 67.2 & 65.6 & 1.207 & 0.511
& 75.8 & 73.8 & \cellcolor{third}0.676 & 0.564
& \cellcolor{third}76.3 & \cellcolor{second}75.0 & 0.502 & \cellcolor{third}0.522
& \cellcolor{second}75.8 & \cellcolor{second}75.6 & 0.389 & \cellcolor{second}0.593 \\
MoMKE~\cite{momke} & MM'24 & \cmark
& 65.2 & 62.8 & 1.200 & 0.474
& 75.7 & \cellcolor{third}75.2 & 0.683 & \cellcolor{third}0.574
& 73.0 & 71.1 & 0.511 & 0.446
& 74.9 & 74.8 & 0.385 & 0.579 \\
LNLN~\cite{lnln} & NeurIPS'24 & \xmark
& 66.2 & 64.0 & 1.182 & 0.477
& 73.8 & 73.5 & 0.703 & 0.543
& \cellcolor{second}76.7 & \cellcolor{third}74.8 & \cellcolor{second}0.479 & \cellcolor{second}0.526
& 75.1 & 74.1 & 0.382 & 0.572 \\
MPLMM~\cite{mplmm} & ACL'24 & \cmark
& \cellcolor{third}68.5 & \cellcolor{second}67.7 & 1.143 & \cellcolor{third}0.520
& 74.8 & 72.6 & 0.688 & 0.567
& 72.7 & 72.3 & 0.495 & 0.500
& 75.1 & 73.8 & 0.380 & \cellcolor{third}0.581 \\
\midrule
LLM-Fusion & -- & \cmark
& 67.7 & 66.9 & \cellcolor{second}1.127 & 0.504
& \cellcolor{third}76.2 & \cellcolor{second}75.5 & \cellcolor{third}0.676 & \cellcolor{second}0.587
& 73.8 & 73.1 & \cellcolor{third}0.484 & 0.496
& 75.5 & \cellcolor{third}75.4 & \cellcolor{second}0.366 & \cellcolor{second}0.593 \\
\rowcolor{oursrow}
CoVeR (ours) & -- & \cmark
& \cellcolor{best}\textbf{69.1} & \cellcolor{best}\textbf{68.9} & \cellcolor{best}\textbf{1.119} & \cellcolor{best}\textbf{0.536}
& \cellcolor{best}\textbf{77.0} & \cellcolor{best}\textbf{76.2} & \cellcolor{best}\textbf{0.659} & \cellcolor{best}\textbf{0.603}
& \cellcolor{best}\textbf{77.2} & \cellcolor{best}\textbf{76.1} & \cellcolor{best}\textbf{0.469} & \cellcolor{best}\textbf{0.530}
& \cellcolor{best}\textbf{76.9} & \cellcolor{best}\textbf{76.7} & \cellcolor{best}\textbf{0.357} & \cellcolor{best}\textbf{0.615} \\
\multicolumn{3}{@{}l}{\quad $\Delta$ vs LLM-Fusion} & +1.4 & +2.0 & -0.008 & +0.032 & +0.8 & +0.7 & -0.017 & +0.016 & +3.4 & +3.0 & -0.015 & +0.034 & +1.4 & +1.3 & -0.009 & +0.022 \\
\bottomrule
\end{tabular}
\vspace{-6pt}
\end{table*}

\subsection{CoVeR: views as trained judges}
\label{sec:cover}
CoVeR keeps the architecture, the inputs and the inference procedure of LLM-driven fusion unchanged and modifies only
how it is trained: each sample is expanded into its full and single-modality views, processed jointly and \emph{with}
gradients; the views are trained as unimodal judges, and their disagreement decides which samples supervise the fused
path.

\para{Counterfactual views as unimodal judges.}
\label{sec:judges}
CoVeR supervises every observed single-modality view with its unimodal label:
\begin{equation}
\Luni=\frac{1}{\sum_{i,j} m_{ij}}\sum_{i}\sum_{j\in\M} m_{ij}\,\ell\big(\hat y^{(j)}_i,\,y^{(j)}_i\big).
\label{eq:uni}
\end{equation}
The labels $y^{(j)}$ are the human unimodal annotations on CH-SIMS/CH-SIMSv2 and the out-of-fold expert predictions of
\cref{sec:conflict} on CMU-MOSI/MOSEI. The mask $m_{ij}$ restricts this loss to observed modalities, so a masked
modality supplies neither an input nor a unimodal target at that visit. Unlike unimodal multi-task heads~\cite{chsims,selfmm}, the judges
are the fusion LLM itself on counterfactual inputs, so their faithfulness shapes the representations the fused path
reads.

\para{Disagreement-gated supervision.}
\label{sec:gate}
CoVeR applies the rule of \cref{eq:gate,eq:mix} during training, with the margin $\tau$, and puts the task loss on the
gated prediction $\tilde y$ instead of $\hat y$:
\begin{equation}
\mathcal L=\lambda_{\mathrm{task}}\big[\ell(\tilde y,y)+\lambda_{\mathrm{uni}}\Luni\big]
+\lambda_{\mathrm{con}}\mathcal L_{\mathrm{con}}+\lambda_{\mathrm{reg}}\mathcal L_{\mathrm{reg}} .
\label{eq:loss}
\end{equation}
Since $\sum_j w_j\in\{0,1\}$, $\partial\ell(\tilde y,y)/\partial\hat y=(1-\sum_j w_j)\,\ell'(\tilde y,y)$: if no judge
disagrees, the fused path is trained as usual; if a judge disagrees, $\tilde y$ is a constant and the fused path receives
no task gradient from that sample, whereas all views keep their unimodal supervision. Why withhold rather than fit?
When the text label differs from the fused target, their losses can pull the shared LoRA parameters in opposite
directions~\cite{pcgrad}. Withholding the fused task gradient preserves unimodal semantic
supervision while fitting the fused path on ungated samples.

\label{sec:deploy}
At test time, \emph{accuracy mode} outputs $\hat y$ of the full view in one forward pass with the architecture and the
inference cost of LLM-Fusion (all main results); the optional \emph{robust mode} applies \cref{eq:gate,eq:mix} with
$\tau_{\mathrm{test}}$, as the override of \cref{sec:override} but with trained judges.

\para{CoVeR improves correlation.}
\Cref{tab:reliability} compares training-time and inference-time corrections. Fixed-threshold discrepancy steering
barely changes ShortErr; test-time override trades consensus accuracy for conflict accuracy.
CoVeR improves Avg Corr by about 2.8/2.0 points on CMU-MOSI/CH-SIMSv2. Its ShortErr values are
71.7/40.5\%, respectively. Consensus accuracy rises from 94.1\% to 95.2\% on CMU-MOSI
and remains at 93.7\% on CH-SIMSv2 to the reported precision.
The two components of CoVeR are ablated on CH-SIMS in \cref{sec:ablation}.

\subsection{Comparison with the state of the art}
\label{sec:sota}
CoVeR achieves the best Acc2, F1 and correlation and the lowest MAE on all four benchmarks (\cref{tab:sota}),
although, like LLM-Fusion, it never reads the features of a missing modality, whereas five of the nine methods use
them as reconstruction or distillation targets; on CH-SIMS it stays ahead of LNLN and DiCMoR also when these are trained with the same
weight averaging (0.530 \vs at most 0.527 Avg Corr). Over LLM-Fusion, which has the same architecture and
inference cost, CoVeR improves the average correlation by approximately +3.2/+1.6/+3.4/+2.2 points and Acc2 by
+1.4/+0.8/+3.4/+1.4 points on CMU-MOSI/MOSEI/CH-SIMS/CH-SIMSv2 (differences of displayed means). Available per-pattern breakdowns and random-missing-rate results are
reported in the supplementary material.

\para{IEMOCAP and MR-Bench-H.}
On IEMOCAP, where the visual channel is informative (a visual-only judge reaches 0.63 test correlation \vs 0.49 for
audio), CoVeR is best on all four metrics against the same nine methods (\cref{tab:iemocap}),
about +1.0 correlation points over LLM-Fusion and +5.4 over the best conventional method (MMIN), and solves
67.4\% of the MR-Bench-A conflict cases against 54.0\% for the backbone. On all 900 clips of MR-Bench-H, CoVeR in
accuracy mode solves 61.7\% of the LS and 93.6\% of the NS clips, versus 58.8\% and
93.1\% for LLM-Fusion, and LS ShortErr falls from 41.2\% to 38.3\% (mean of three seeds;
\cref{tab:s-mrbh}). The best re-run method on LS clips, MMIN, reaches 60.7\% LS and 91.9\% NS accuracy. Robust mode reaches 63.0\% LS but 84.8\% NS accuracy, trading NS
accuracy for LS accuracy. Against an EmoMM-inspired state-steering adaptation of LLM-Fusion (61.2\% LS,
91.2\% NS), accuracy-mode CoVeR gains about 0.5 LS points (from the displayed means). Among the LS clips a model solves from audio and video
alone, adding the text makes 32.2\% wrong for LLM-Fusion but 30.1\% for CoVeR, direct evidence of
less text-induced overriding.

\begin{table}[tb]
\centering
\caption{\textbf{Supervision ablation on CH-SIMS.} Accuracy mode; Avg Corr ($\times100$) across seven
patterns, conflict Acc2 and ShortErr (SE) on MR-Bench-A (\%). Mean over five seeds; the LLM-Fusion row is the
three-seed reference of \cref{tab:sota,tab:bench}.}
\label{tab:ablation}
\small
\setlength{\tabcolsep}{3pt}
\begin{tabular}{@{}lrrr@{}}
\toprule
Variant & Avg Corr\up & Conf.\up & SE\dn\\
\midrule
LLM-Fusion & 49.6 & 38.7 & 61.3\\
+ unimodal loss, no gate & 52.2 & 41.3 & 58.7\\
\rowcolor{oursrow}
+ unimodal loss + gate (CoVeR) & 55.1 & 47.7 & 52.3\\
\bottomrule
\end{tabular}
\vspace{-6pt}
\end{table}

\subsection{Ablation study}
\label{sec:ablation}
\Cref{tab:ablation} separates the two components of CoVeR on CH-SIMS, where human unimodal labels supervise the
judges. Unimodal supervision without the gate improves Avg Corr by 2.6 points and conflict accuracy by 2.6 points
over LLM-Fusion. Adding the disagreement gate gains a further 2.9 Corr points (from displayed means) and 6.4 conflict points. This isolates
the benefit of suppressing incompatible fused supervision beyond learning unimodal semantics. Under permanent missingness, where every training and validation sample keeps one availability pattern for the whole run, CoVeR reaches 49.5/58.8/50.0/59.8 Avg Corr ($\times100$) on CMU-MOSI/MOSEI/CH-SIMS/CH-SIMSv2 against 47.8/56.0/46.6/57.9 for LLM-Fusion (\cref{tab:s-fixed}).

\para{Collapse and cost.}
CoVeR weakens the collapse: the late-layer text reliance on conflict cases drops from 0.81 to 0.74 (CMU-MOSI) and
0.77 to 0.72 (CH-SIMS), and the AUROC of the judges' disagreement rises from 0.72 to 0.86 on CMU-MOSI (\cref{fig:diag}d).
Accuracy-mode inference costs the same as LLM-Fusion (4.0\,ms per clip on an RTX 4090); a training epoch takes
5.7$\times$ as long and robust mode 6$\times$ the latency.
\section{Related Work}
\label{sec:related}

\para{Incomplete multimodal learning.}
Missing modalities are reconstructed with autoencoders~\cite{mmin}, transformers~\cite{tfrnet},
graphs~\cite{gcnet}, diffusion~\cite{imder,dicmor} or masked modelling~\cite{actionmae}; other approaches
distil complete-modality knowledge~\cite{corrkd,pcd,momke,mcmoe}, learn missing-aware
prompts~\cite{lee2023missingprompt,mplmm}, or retrieve and translate missing content~\cite{missrag,tamml}.
Modality imbalance~\cite{wang2020makes,huang2022modality,du2023unimodal} motivates gradient
modulation~\cite{ogmge,pmr,agm}, objective balancing~\cite{mmpareto} and uncertainty weighting~\cite{tmc,mmdynamics,qmf}.
In sentiment analysis, text dominates~\cite{almt,mult,misa,lnln}; unimodal labels support multi-task
learning~\cite{chsims,selfmm}. CoVeR uses these labels to train counterfactual views of the fusion LLM and
withholds fused supervision when the views disagree.

\para{Text bias and modality conflict.}
\emph{Words or Vision}~\cite{wordsorvision} studies text bias in VLMs under visual--textual conflicts,
including model scale, prompts and token order, and explores supervised fine-tuning with text augmentation.
Our distinction is the training objective and setting: we study feature-input LLM fusion under natural sentiment
conflict and gate fused supervision when it conflicts with unimodal semantics.
EmoMM~\cite{emomm} benchmarks aligned, conflicting and missing emotion cues, diagnoses Video Contribution
Collapse, and proposes conflict-aware head-level attention steering (CHASE) at inference time.
MR-Bench-H instead conditions its LS challenge set on text-reader failures. Our diagnosis concerns late-decoder
alignment with the text view; our correction trains unimodal judges and gates the fused task gradient.
Our EmoMM-inspired comparator steers fusion-token states rather than reproducing CHASE head-level attention
(\cref{tab:s-mrbh}).
Together with view-based inference~\cite{cfvqa,vcd,iti,caa} and shortcut mitigation~\cite{geirhos2020shortcut,rubi,lmh,css},
these studies motivate our focus on resolving conflicting supervision during training.
\section{Conclusion}
\label{sec:conclusion}

MR-Bench diagnoses conflicting modality evidence. Its human-validated release has 620 LS and 280 NS clips;
every model is scored on all 900 clips. Text inversion alone reaches 94.4\% LS accuracy,
making joint LS/NS and paired-view evaluation essential. CoVeR achieves state-of-the-art results on five benchmarks.
It reaches 61.7\% LS / 93.6\% NS accuracy without extra inference cost; robust mode reaches 63.0\% / 84.8\%,
leaving a trade-off between the two shortcut types. Paired views show less text-induced overriding:
the AV-correct-to-TAV-wrong rate falls from 32.2\% to 30.1\%.
\emph{Limitations}: the CMU-MOSI/MOSEI conflict sets of MR-Bench-A are model-judged and small, MR-Bench-H holds only model-detectable shortcuts, and CoVeR needs unimodal labels and a known dominant modality.

\clearpage
\section*{LLM Usage Statement}
OpenAI Codex assisted with language editing, LaTeX formatting, dataset-release documentation and supporting scripts. Experimental measurements and human-validation statistics reported here were
supplied by the authors, not generated by the tool. The authors are responsible for reviewing the assisted text and
code and for the final content of this work.
{
    \small
    \bibliographystyle{ieeenat_fullname}
    \bibliography{main}
}

\clearpage
\maketitlesupplementary
\setcounter{section}{0}
\renewcommand{\theHsection}{supp.\arabic{section}}
\setcounter{table}{0}
\setcounter{figure}{0}
\renewcommand{\thesection}{\Alph{section}}
\renewcommand{\thetable}{S\arabic{table}}
\renewcommand{\thefigure}{S\arabic{figure}}

\noindent This supplement gives the details needed to reproduce the paper (\cref{sec:s-impl}), the construction of
\bench with sample clips (\cref{sec:s-bench}), the complete \benchH results on all 900 clips (\cref{sec:s-mrbh}), the
analyses referenced in the main paper (\cref{sec:s-diag}), additional results (\cref{sec:s-results}) and the
permanent-missingness protocol (\cref{sec:s-fixed}). Numbers without the prefix ``S'' refer to the main paper.

\section{Implementation details}
\label{sec:s-impl}

\para{Input sequence.}
Each modality sequence is resampled to $T=50$ steps by adaptive average pooling over its valid (unpadded) length
(audio and visual features are z-scored with training-split statistics) and projected into the LLM embedding space by a modality-specific projection (two $1{\times}1$ convolutions with
normalisation, ReLU and dropout). The prompt $P(\mathbf p)$ is: \emph{``Given the available
\{avail.\} features from a video clip. The \{miss.\} modality is missing. Based on the available modalities, please infer
and reconstruct the useful latent representations for the missing \{miss.\} modalities at the designated positions.
Then integrate and enhance all multimodal features for multimodal sentiment analysis. Output the fused
multi-dimensional feature representations at the designated feature dimension positions:''}. The eight prompt variants
(one per availability pattern, including the empty one used by $\V_\varnothing$) are tokenised once and left-padded to a
common length, so that every view has an identical token layout.

\para{Metrics.}
Acc2 and F1 exclude neutral labels; F1 is scikit-learn's weighted F1. Averages over the seven patterns (Avg) and
over the seven missing rates (Rate) are arithmetic means, except for the correlation, which is averaged through
Fisher's $z$-transform: $\bar r_z=\tanh\!\big(K^{-1}\sum_{k=1}^{K}\operatorname{arctanh}(r_k)\big)$,
with $K=7$. Tables reporting Corr $\times100$ apply this scaling after aggregation.

\para{Datasets and hyper-parameter selection (\cref{sec:setup}).}
Train/validation/test clips: CMU-MOSI 1,284/229/686, CMU-MOSEI 16,326/1,871/4,659, CH-SIMS 1,368/456/457, CH-SIMSv2
2,722/647/1,034 and IEMOCAP 6,740/617/2,170 (Sessions 1--4 split by dialogue, Session 5 test). We use Qwen3-0.6B~\cite{qwen3} with LoRA~\cite{lora} ($\alpha=2r$) on the attention projections, trainable
projections, special tokens and heads, and $K=4$ fusion tokens; $\lambda_{\mathrm{task}}=10$ and
$\lambda_{\mathrm{con}}=\lambda_{\mathrm{reg}}=1$ for \base and \ours. We train with AdamW~\cite{adamw} (cosine schedule),
batch 32, bf16 and activation checkpointing~\cite{chen2016ckpt}, with a default budget of 30 epochs (15 on CMU-MOSEI), on one 24\,GB GPU.
Hyper-parameters are selected per dataset on the validation split only, over two seeds: $\tau\in\{0.2,0.3,0.4\}$,
$\lambda_{\mathrm{uni}}\in\{0.5,1,2\}$ and rank $r\in\{16,32\}$, then learning rate, epochs and an exponential moving
average of the weights (decay $\in\{0,0.99,0.995\}$), on CH-SIMS also the regression loss (MSE or L1); CMU-MOSEI reuses
the CMU-MOSI setting. The selection for \ours is $\tau=0.4$ (labels normalised to $[0,1]$, $c=0.5$), $\lambda_{\mathrm{uni}}=2$,
the MSE loss and lr $2\times10^{-4}$ everywhere, $r=16$ except on CH-SIMS ($r=32$), and weight averaging (0.99) except on
CH-SIMSv2. \base receives the same search (with the MSE loss); its selection is $r=32$ and lr $2\times10^{-4}$ on
CMU-MOSI and CMU-MOSEI, $r=16$ and lr $10^{-4}$ on CH-SIMS (50 epochs) and CH-SIMSv2, and weight averaging (0.99) on
all four. LNLN and DiCMoR are also re-run with weight averaging
on CH-SIMS. The analyses in \cref{tab:override,tab:reliability,tab:ablation} use the default recipe of the backbone for
every row ($\tau=0.3$, $\lambda_{\mathrm{uni}}=1$ for \ours). Metrics: means over three seeds for the comparisons with
the state of the art and the diagnosis of \cref{sec:findings}, and over five seeds for the analyses of \ours. GCNet~\cite{gcnet}
needs whole conversations as graphs and we could not obtain the code of CorrKD~\cite{corrkd}, so neither is re-run.

\para{Views, labels and the gate.}
For a mini-batch of $B$ samples, \ours processes the full view and the three single-modality views ($4B$ sequences) in
one forward pass with gradients and activation checkpointing~\cite{chen2016ckpt}; a view whose modality is not observed
is masked out of $\Luni$ and of the gate. Labels and unimodal labels are normalised to $[0,1]$ with the label range of
the dataset, so the neutral point is $c=0.5$.

\para{Unimodal experts.}
For CMU-MOSI/MOSEI, each expert is a LayerNorm, a bidirectional LSTM (hidden size 128) with mean pooling and a two-layer
MLP head (dropout 0.3), trained with AdamW (learning rate $10^{-3}$, weight decay $10^{-3}$, batch 64) for 40 epochs on
one modality. Three seeds are averaged. On the training split we use five-fold out-of-fold predictions.

\para{Features extracted by us (IEMOCAP, MELD, MUStARD++).}
IEMOCAP~\cite{iemocap} has no MMSA features, so we extract them: text = BERT-base last hidden states of the transcript
($\le$50 word pieces, the MMSA recipe); audio = openSMILE ComParE\_2016 low-level descriptors (65-d, 100\,fps) of the
utterance; visual = the CLS state and the 7 expression logits of a ViT fine-tuned on FER2013 (775-d) on the speaker's face,
detected by MTCNN at 5\,fps in the dialogue video. Both actors are in frame; the female actor's side is read off five
frames per dialogue by Qwen2.5-VL-7B (asked twice, on the frame and its mirror image; 148 of 151 dialogues fully
consistent), and the utterance id gives the speaker's gender. Every stream is pooled to $T=50$. The score is the averaged
valence (1--5) minus 3, so polarity is its sign and neutral utterances are excluded from Acc2/F1. Sessions 1--4 form the
training set with 10\% of their dialogues as validation, Session 5 is the test set, and the 512 \benchH candidates of
Sessions 1--4 are held out of training. Unimodal judgments for \benchA and for the judges' supervision come from
out-of-fold experts as on CMU-MOSI/MOSEI (test correlation: text 0.65, audio 0.49, visual 0.63). \base and \ours are trained
for 20 epochs; the remaining settings are those of CMU-MOSI, including weight averaging.

MELD~\cite{meld} and MUStARD++~\cite{mustardpp}, which are needed to score fusion models on their \benchH clips
(\cref{sec:s-mrbh}), use the same three extractors; their clips do not identify the speaker, so the largest confident
face of each frame is embedded. MELD: the official training split minus 10\% of its dialogues (validation) is used for
training, with the sentiment label in $\{-1,0,1\}$ as the score; all \benchH clips come from the official development
and test splits. MUStARD++: every \benchH candidate is held out and the remaining utterances are split 80/10/10, with
the score $(\text{valence}-5)/4$. Unimodal judgments come from out-of-fold experts. LLM-Fusion and CoVeR use the
IEMOCAP recipe, with 15 epochs on MELD and 30 on MUStARD++; the re-run methods keep their default budgets (30 epochs;
MMIN 40 and 80), except DiCMoR and IMDer on MELD (15 epochs).

\para{Common protocol for the baselines.}
Every baseline is ported from its official repository into our training loop, which draws one of the seven availability
patterns per sample and epoch, zero-fills missing modalities and passes the pattern to methods that use it (MMIN,
TFR-Net, IMDer, DiCMoR, LNLN, MPLMM, MoMKE). Inputs are the frozen BERT, audio and visual features used by \ours, resampled
to 50 steps; where the official code fine-tunes BERT, the BERT features enter where its last hidden states did. Official
hyper-parameters (optimizer, learning rate, batch size, loss weights, depths, dropouts, gradient clipping and
accumulation) are taken from the authors' configurations for CMU-MOSI, CMU-MOSEI and CH-SIMS (CH-SIMS values are used for
CH-SIMSv2). All methods train for the same budget as \ours (30 epochs, 15 on CMU-MOSEI; MMIN see below) and are tested identically. We do not copy published numbers: protocols differ
(token-level \vs whole-modality missingness), some were computed with an earlier MMSA F1 implementation that overestimates
F1 on CH-SIMS (corrected in the LNLN repository); all F1 values here use scikit-learn's weighted F1 with the true
labels as reference. Methods whose objectives use complete modalities as
reconstruction, flow or distillation targets receive the unmasked features for these targets only (T-Miss \xmark{} in
\cref{tab:sota}); \base and \ours never read the features of a missing modality. In every method the availability
pattern is re-drawn at each visit of a training sample from the seven non-empty patterns, the complete one included
(probability $1/7$ each), so the masking is dynamic: over 30 epochs a sample is seen complete about four times, and each of its modalities is
observed in $4/7$ and hidden in $3/7$ of its visits. Training sets with permanently missing modalities, where a
modality of a sample is never observed, are evaluated separately in \cref{sec:s-fixed}. \Cref{tab:s-baselines} lists the remaining deviations.

\begin{table}[t]
\centering
\caption{\textbf{Baseline ports.} Complete-modality targets used in training, and deviations from the official code
beyond the common protocol.}
\label{tab:s-baselines}
\scriptsize
\setlength{\tabcolsep}{2.4pt}
\begin{tabular}{@{}lcp{5.2cm}@{}}
\toprule
Method & Targets & Deviations\\
\midrule
MulT~\cite{mult}      & --      & none (zero imputation)\\
MISA~\cite{misa}      & --      & utterance vectors by mean pooling (no padding)\\
MMIN~\cite{mmin}      & \cmark  & full-modality teacher (first 25\% of epochs) then imagination; official IEMOCAP settings; 80 epochs (22 on CMU-MOSEI) and batch 32--128 to match the official number of updates, since 30 epochs leave the teacher under-trained\\
TFR-Net~\cite{tfrnet} & \cmark  & missing positions from the pattern (whole modality) instead of random frames\\
IMDer~\cite{imder}    & \cmark  & score-based recovery per sample from the observed modalities (official 100 sampling steps); trained from scratch instead of from the official complete-data checkpoint\\
DiCMoR~\cite{dicmor}  & \cmark  & flow-based recovery per sample from the observed modalities; trained from scratch instead of from the official complete-data checkpoint\\
MoMKE~\cite{momke}    & --      & expert stage (first half of the epochs, each expert trained on samples observing its modality) then soft-routed mixture; experts attend over the frames of one clip instead of the utterances of one video\\
LNLN~\cite{lnln}      & \cmark  & completeness label from the pattern; warm-up/cosine schedule rescaled to 30 epochs\\
MPLMM~\cite{mplmm}    & --      & trained end to end (no complete-data pre-training); missing-type prompts initialised
                                   $\mathcal N(0,0.02^2)$, since the official zero initialisation receives no gradient\\
\bottomrule
\end{tabular}
\end{table}

\para{MLLM evaluation.}
Each clip is presented to an MLLM under every view it supports: full clip (video with its audio track), transcript only,
muted video, audio only, and, for vision-only models, video with the transcript. The question is ``Is the overall
sentiment expressed by the speaker in this clip positive or negative? Answer with exactly one word: positive or
negative.'', prefixed by the transcript for transcript views; Chinese clips are prompted in English. Decoding is greedy
with at most six new tokens; the first occurrence of \emph{positive} or \emph{negative} is the answer, and any other
answer counts as wrong. Neutral or invalid answers are not counted as text-following errors. Qwen2.5-Omni samples video at 2\,fps with at most
$360\times420$ pixels per frame and interleaves the audio (\texttt{use\_audio\_in\_video}); MiniCPM-o~2.6 receives one
frame and one second of 16\,kHz audio per unit (at least four units); the video-only models receive 16 uniformly sampled
frames. All models run in bf16 on one GPU with the official chat templates.

\section{\bench: construction and sample clips}
\label{sec:s-bench}

\para{Case rules.}
With polarity $p(\cdot)=\sgn(\cdot)$ and confidence threshold $\kappa$, among samples with $p(y)\neq0$:
\emph{consensus} if $p(u^{(t)})=p(u^{(a)})=p(u^{(v)})=p(y)$; \emph{conflict} if $p(u^{(t)})\notin\{0,p(y)\}$,
$|u^{(t)}|\ge\kappa$, and $\exists j\in\{a,v\}$ with $p(u^{(j)})=p(y)$, $|u^{(j)}|\ge\kappa$; \emph{weak-modality
conflict} if $p(u^{(t)})=p(y)$ and $\exists j\in\{a,v\}$ with $p(u^{(j)})\notin\{0,p(y)\}$, $|u^{(j)}|\ge\kappa$;
\emph{other} otherwise. $\kappa=0.2$ for CH-SIMS/CH-SIMSv2 and $0.5$ for CMU-MOSI/MOSEI.

\para{Expert quality.}
On the test splits, the text/audio/visual experts reach Corr 0.68/0.18/0.09 on CMU-MOSI and 0.75/0.24/0.28 on CMU-MOSEI,
mirroring the dominance of text in these datasets. The mined conflict sets are therefore conservative: a case enters only
if a weak expert is confidently right where the strong text expert is confidently wrong.

\begin{table}[t]
\centering
\caption{\textbf{\bench counts per split} (train / validation / test). ``Weak'': weak-modality conflict.}
\label{tab:s-bench}
\scriptsize
\setlength{\tabcolsep}{3.5pt}
\begin{tabular}{@{}lrrrr@{}}
\toprule
Dataset & Samples & Consensus & Conflict & Weak\\
\midrule
CMU-MOSI  & 1284/229/686     & 428/92/187     & 55/9/36  & 297/55/186\\
CMU-MOSEI & 16326/1871/4659  & 6387/675/1669  & 68/4/20  & 675/83/265\\
CH-SIMS   & 1368/456/457     & 621/211/221    & 105/37/31 & 99/38/32\\
CH-SIMSv2 & 2722/647/1034    & 1423/326/457   & 148/55/73 & 146/42/54\\
IEMOCAP   & 6740/617/2170 & 4660/312/951 & 50/14/87 & 109/34/200\\
\bottomrule
\end{tabular}
\end{table}

\begin{table}[t]
\centering
\caption{\textbf{MR-Bench-A and the conflict patterns of LLM-Fusion.} Test-split consensus (Cons.) and conflict
(Conf.) counts and, with all modalities observed, Acc2 on both case types, ShortErr (SE) and conflict Acc2 of the same
model without text (w/o $t$) and of its visual-only view (by $\V_v$), in \%; mean over 3 seeds.}
\label{tab:bench}
\scriptsize
\setlength{\tabcolsep}{2.1pt}
\begin{tabular}{@{}ll rr ccccc@{}}
\toprule
\multirow[c]{2}{*}[-2.4pt]{Dataset} &\multirow[c]{2}{*}[-2.4pt]{Judgments} & \multicolumn{2}{c}{\#Test} & \multicolumn{5}{c}{LLM-Fusion (\%)} \\
\cmidrule(lr){3-4}\cmidrule(l){5-9}
 & & Cons. & Conf. & Cons.\up & Conf.\up & SE\dn & w/o $t$\up & by $\V_v$\up \\
\midrule
CMU-MOSI  & OOF experts & 187  & 36 & 94.7&22.2&77.8&63.0&54.6\\
CMU-MOSEI & OOF experts & 1669 & 20 & 97.8&16.7&83.3&86.7&78.3\\
CH-SIMS   & human       & 221  & 31 & 89.6&38.7&61.3&68.8&69.9\\
CH-SIMSv2 & human       & 457  & 73 & 93.7&59.4&40.6&83.1&84.0\\
\bottomrule
\end{tabular}
\end{table}

\para{\benchH, stage 1 (candidate mining).}
A clip with a non-neutral label is an LS candidate if its text judgment has the opposite polarity with confidence at
least $\kappa_t$, and an NS candidate if the text judgment agrees with the label while the audio or visual judgment has
the opposite polarity with confidence at least $\kappa_{a}$ or $\kappa_{v}$; $(\kappa_t,\kappa_a,\kappa_v)=(0.5,0.3,0.3)$
on CMU-MOSI/MOSEI and $(0.2,0.2,0.2)$ on CH-SIMS/v2. Judgments are those of \benchA (validation/test) or of the
out-of-fold experts (training splits). On MUStARD++~\cite{mustardpp}, a clip is an LS candidate if its explicit emotion
has the opposite polarity of its implicit emotion and the valence rating (1--9) confirms the implicit polarity
($\ge6$ positive, $\le4$ negative). On MELD~\cite{meld}, the text judgment is the text-only view of \base trained on
CMU-MOSEI, applied to BERT features extracted with the MMSA pipeline (threshold 0.5; Acc2 64.8\% on the non-neutral
MELD utterances). IEMOCAP~\cite{iemocap} is handled like MELD; its reference polarity is the averaged valence
($\ge3.5$ positive, $\le2.5$ negative) when the emotion category agrees (happy/excited \vs angry/sad/frustrated/fearful/
disgusted), non-verbal annotations such as [LAUGHTER] are removed from the transcripts, and utterance clips are cut
from the dialog videos, which show both actors (annotators judge the speaker).

\para{Stage 2 (reader-panel screening).}
Four instruction-tuned text LLMs from three families (Qwen2.5-7B, Qwen3-8B, Yi-1.5-9B, Phi-3.5-mini) answer the
binary question from the transcript alone, with the protocol of the MLLM survey. An LS candidate is kept if at least
three of them answer the misleading polarity, \ie the words mislead strong readers; NS candidates and controls are kept
if at least three answer the label polarity, \ie the words carry the evidence (whether the face misleads is decided by
the annotators). The panel is independent of the fusion models. All reported MR-Bench-H evaluations use the same
900 released clips, including the evaluation of the text-only panel member. A first version screened with the restricted views of \base only;
it disadvantaged models sharing that backbone and kept English candidates whose words do not mislead strong readers
(a 7B text LLM solved 45\% of them), which motivated the panel. Of 2,040 candidates, 1,337 pass (798 LS of 1,474, 363
NS of 380, 176 controls of 186); of the 2,593 reserve candidates, 1,563 pass.

\para{Stage 3 (blind validation).}
Annotators receive shuffled clips without labels (guide released with the benchmark; about 50\,s per clip) and answer,
in this order: Q2 sentiment from the transcript alone, Q5 from the muted video, Q1 from the full clip (negative, neutral,
positive), Q3 the decisive channel (words, voice, face, mixed) and Q4 ambiguity. With three annotators per clip, an LS
clip is kept if the majority gives Q1 = label, Q2 = misleading polarity, Q3 $\in$ \{voice, face, mixed\} and Q4 = no; an
NS clip if Q1 = label, Q5 = misleading polarity, Q3 $\in$ \{words, voice, mixed\} and Q4 = no. Every retained LS and NS
clip meets these criteria, so majority-vote accuracy on the released clips is fixed by the rule and is not reported as a
result. Annotator agreement is Fleiss' $\kappa=0.75$. Control clips are excluded from the release and are not selected
by these criteria; the majority answer matches the label on 91.0\% of them, which flags careless annotators.

\begin{table}[t]
\centering
\caption{\textbf{MR-Bench-H per source}: candidates after stage 1, items passing the reader panel (stage 2), and items
retained after human validation (stage 3). Validated columns report the final retained counts in the release package.
Controls are tracked separately and excluded from the released test set. LS / NS / controls.}
\label{tab:s-benchh}
\scriptsize
\setlength{\tabcolsep}{2.0pt}
\begin{tabular}{@{}ll ccc ccc ccc@{}}
\toprule
\multirow[c]{2}{*}[-2.4pt]{Source} &\multirow[c]{2}{*}[-2.4pt]{Lang.} & \multicolumn{3}{c}{Candidates} & \multicolumn{3}{c}{Reader panel} & \multicolumn{3}{c}{Validated} \\
\cmidrule(lr){3-5}\cmidrule(lr){6-8}\cmidrule(l){9-11}
 & & LS & NS & Ctrl & LS & NS & Ctrl & LS & NS & Ctrl\\
\midrule
CMU-MOSI~\cite{mosi}        & EN & 83  & 135 & 22 & 23  & 130 & 21 & 14 & 119 & 0 \\
CMU-MOSEI~\cite{mosei}      & EN & 139 & 177 & 32 & 53  & 175 & 28 & 23 & 119 & 0 \\
CH-SIMS~\cite{chsims}       & ZH & 48  & 26  & 7  & 38  & 22  & 7  & 24 & 16 & 0 \\
CH-SIMSv2~\cite{chsimsv2}   & ZH & 96  & 42  & 14 & 74  & 36  & 14 & 49 & 26 & 0 \\
MUStARD++~\cite{mustardpp}  & EN & 101 & --  & 10 & 69  & --  & 10 & 53 & -- & 0 \\
MELD~\cite{meld}            & EN & 407 & --  & 41 & 176 & --  & 39 & 176 & -- & 0 \\
IEMOCAP~\cite{iemocap}      & EN & 600 & --  & 60 & 365 & --  & 57 & 281 & -- & 0 \\
\midrule
Total                       &    & 1,474 & 380 & 186 & 798 & 363 & 176 & 620 & 280 & 0 \\
\bottomrule
\end{tabular}
\end{table}

\begin{figure*}[p]
\centering
\includegraphics[width=\textwidth,height=0.90\textheight,keepaspectratio]{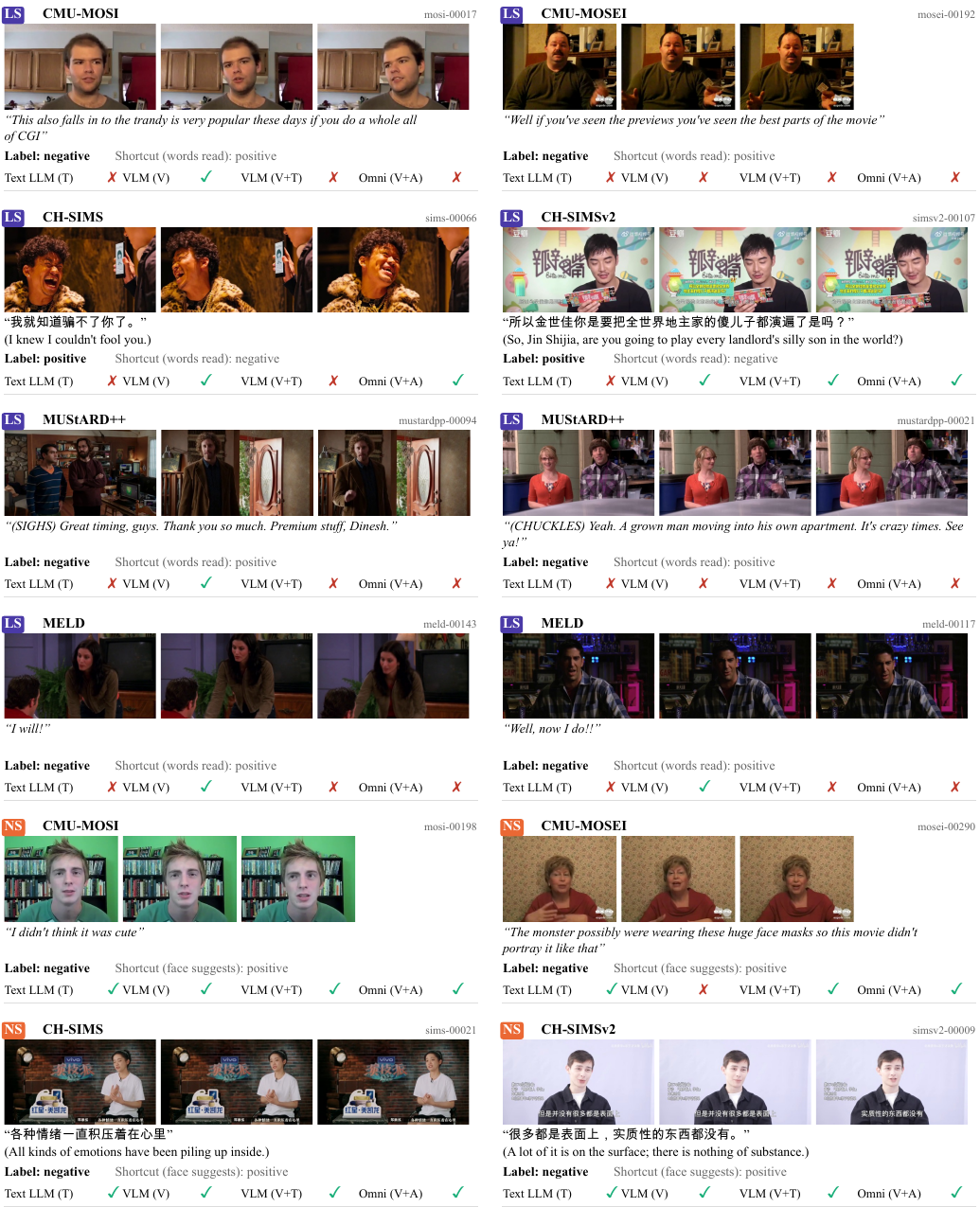}
\caption{\textbf{Sample clips of \benchH} (frames at 20/50/80\% of each clip; Chinese transcripts with an English
gloss). LS: the words suggest the polarity opposite to the label; NS: the face does. Marks show whether a reader
answers the label: Qwen2.5-7B on the transcript (T), InternVL3-8B on the muted video (V) and on the video with its
transcript (V+T), Qwen2.5-Omni-7B on the full clip (V+A).}
\label{fig:s-samples}
\end{figure*}

\para{Sample clips.}
\Cref{fig:s-samples} shows twelve clips of \benchH with their transcripts, labels, shortcut polarities and the recorded
answers of four zero-shot readers. They are not hand-picked for the outcome: for each source and type we take the
first clips of a seeded random order that show the speaker clearly, skipping letterboxed frames, profanity and
politically sensitive content. IEMOCAP clips are part of the benchmark but are not reproduced here because of its
redistribution terms. On the LS clips the text reader follows the words, the vision--language model often reads the
muted video correctly and loses this answer once the transcript is added; on the NS clips the words carry the label
and the muted video alone is the unreliable view.

\afterpage{\clearpage}

\para{Licences and intended use.}
\bench adds derived annotations keyed by the original sample identifiers and inherits the licences and annotation
biases of its sources. The diagnosis tools are intended for auditing models; we do not recommend deploying affect
recognition in high-stakes decisions about individuals.

\section{\benchH results on all 900 clips}
\label{sec:s-mrbh}

\para{Protocol.}
Every model is scored on all 620 LS and 280 NS clips of the release. Majority-vote accuracy of the annotators is fixed by the retention rule in \cref{sec:s-bench} and is omitted from \cref{fig:results,tab:s-mrbh}. An answer that is not a valid polarity counts as
wrong, and \shorterr is the share of LS clips answered with the misleading polarity of the words. Zero-shot MLLMs take
the raw clips (\cref{sec:s-impl}). A fusion model reads features, so each clip is scored, with all modalities observed,
by the model trained on the training split of its source corpus (the four MMSA datasets, IEMOCAP, MELD and MUStARD++);
no benchmark clip is used to update a model. Fusion results are the mean and standard deviation over three seeds of the
accuracy pooled over the seven sources. With 620 LS clips the 95\% Wilson interval of an accuracy is at most $\pm3.9$
points, against $\pm9.2$ points on the 110 LS clips of the four MMSA sources alone.

\begin{table}[t]
\centering
\caption{\textbf{\benchH on all 900 clips.} Accuracy (\%) on the 620 LS and 280 NS clips, 95\% Wilson interval of the LS accuracy and LS \shorterr (SE, \%). V+A: clip with its audio track, V: muted video, V+T: muted video with the transcript, A: audio, T: transcript. Text inversion flips the valid answers of Qwen2.5-7B; always ``negative'' is the constant answer.}
\label{tab:s-mrbh}
\scriptsize
\setlength{\tabcolsep}{2.4pt}
\begin{tabular}{@{}llcccc@{}}
\toprule
Model & Input & LS\up & 95\% CI & NS\up & SE\dn\\
\midrule
\multicolumn{6}{@{}l}{\textit{Zero-shot MLLMs}}\\
Qwen2.5-Omni-7B & V+A & 28.7 & [25.3, 32.4] & 94.6 & 71.3\\
 & A & 21.8 & [18.7, 25.2] & 97.9 & 78.2\\
 & V & 53.9 & [49.9, 57.8] & 33.2 & 46.1\\
 & T & 3.1 & [2.0, 4.7] & 99.3 & 96.8\\
Qwen2.5-Omni-3B & V+A & 25.8 & [22.5, 29.4] & 92.5 & 74.2\\
 & A & 17.6 & [14.8, 20.8] & 94.6 & 82.4\\
 & V & 45.2 & [41.3, 49.1] & 26.1 & 54.8\\
 & T & 4.5 & [3.1, 6.4] & 98.6 & 95.5\\
MiniCPM-o 2.6 & V+A & 30.2 & [26.7, 33.9] & 89.6 & 69.7\\
 & A & 36.8 & [33.1, 40.6] & 93.9 & 62.4\\
 & V & 39.8 & [36.1, 43.7] & 22.1 & 59.2\\
 & T & 3.2 & [2.1, 4.9] & 99.6 & 96.8\\
Qwen2-Audio-7B & A & 32.7 & [29.2, 36.5] & 90.4 & 66.9\\
 & T & 9.5 & [7.4, 12.1] & 98.2 & 89.2\\
InternVL3-8B & V & 67.7 & [64.0, 71.3] & 60.7 & 31.5\\
 & V+T & 21.6 & [18.6, 25.0] & 98.9 & 77.6\\
 & T & 2.6 & [1.6, 4.2] & 100.0 & 97.3\\
LLaVA-OV-7B & V & 68.9 & [65.1, 72.4] & 40.7 & 30.3\\
 & V+T & 15.8 & [13.1, 18.9] & 98.2 & 83.4\\
 & T & 11.6 & [9.3, 14.4] & 99.3 & 88.4\\
Qwen2.5-VL-7B & V & 61.1 & [57.2, 64.9] & 42.5 & 38.1\\
 & V+T & 13.2 & [10.8, 16.1] & 99.3 & 86.0\\
 & T & 3.1 & [2.0, 4.7] & 99.3 & 96.6\\
Qwen2.5-7B & T & 5.5 & [4.0, 7.6] & 99.3 & 94.4\\
Text inversion & T & 94.4 & [92.3, 95.9] & 0.7 & 5.5\\
Always ``negative'' & -- & 61.9 & -- & 83.6 & 38.1\\
\midrule\multicolumn{6}{@{}l}{\textit{Fusion models} (mean$\pm$std over three seeds)}\\
\rowcolor{oursrow}CoVeR, robust & $t,a,v$ & 63.0$\pm$2.0 & -- & 84.8$\pm$0.7 & 37.0\\
\rowcolor{oursrow}CoVeR, accuracy & $t,a,v$ & 61.7$\pm$2.8 & -- & 93.6$\pm$1.3 & 38.3\\
MMIN & $t,a,v$ & 60.7$\pm$1.8 & -- & 91.9$\pm$1.8 & 39.3\\
MulT & $t,a,v$ & 60.4$\pm$3.0 & -- & 91.3$\pm$1.2 & 39.6\\
MoMKE & $t,a,v$ & 59.4$\pm$0.7 & -- & 90.2$\pm$3.1 & 40.6\\
LLM-Fusion & $t,a,v$ & 58.8$\pm$0.8 & -- & 93.1$\pm$0.7 & 41.2\\
\quad + EmoMM-style state steering & $t,a,v$ & 61.2$\pm$1.2 & -- & 91.2$\pm$1.8 & 38.8\\
TFR-Net & $t,a,v$ & 55.4$\pm$4.9 & -- & 90.6$\pm$2.0 & 44.6\\
IMDer & $t,a,v$ & 54.6$\pm$1.5 & -- & 90.7$\pm$2.5 & 45.4\\
MPLMM & $t,a,v$ & 54.3$\pm$1.2 & -- & 91.2$\pm$2.4 & 45.7\\
LNLN & $t,a,v$ & 53.7$\pm$2.6 & -- & 91.1$\pm$1.7 & 46.3\\
MISA & $t,a,v$ & 53.2$\pm$3.0 & -- & 91.4$\pm$0.8 & 46.8\\
DiCMoR & $t,a,v$ & 52.7$\pm$3.2 & -- & 93.8$\pm$0.9 & 47.3\\
\bottomrule
\end{tabular}
\end{table}

\begin{table}[t]
\centering
\caption{\textbf{LS accuracy per source} (\%; number of LS clips in parentheses). Fusion models are trained on the training split of each source; $t$: their text-only view. Mean over three seeds for the fusion models.}
\label{tab:s-mrbh-source}
\scriptsize
\setlength{\tabcolsep}{1.5pt}
\begin{tabular}{@{}llccccccc@{}}
\toprule
Model & Input & \makecell{MOSI\\(14)} & \makecell{MOSEI\\(23)} & \makecell{SIMS\\(24)} & \makecell{SIMSv2\\(49)} & \makecell{MUSt.\\(53)} & \makecell{MELD\\(176)} & \makecell{IEMO.\\(281)}\\
\midrule
Qwen2.5-7B & T & 7 & 4 & 0 & 2 & 6 & 7 & 5\\
InternVL3-8B & V & 71 & 70 & 58 & 39 & 55 & 72 & 73\\
InternVL3-8B & V+T & 14 & 30 & 25 & 31 & 6 & 23 & 22\\
Qwen2.5-Omni-7B & V+A & 29 & 26 & 38 & 43 & 4 & 27 & 31\\
Always ``negative'' & -- & 57 & 48 & 50 & 35 & 100 & 79 & 51\\
\midrule
LLM-Fusion & $t$ & 5 & 7 & 8 & 12 & 83 & 35 & 56\\
LLM-Fusion & $a,v$ & 43 & 51 & 64 & 81 & 99 & 77 & 82\\
LLM-Fusion & $t,a,v$ & 7 & 9 & 15 & 43 & 88 & 46 & 75\\
\bottomrule
\end{tabular}
\end{table}

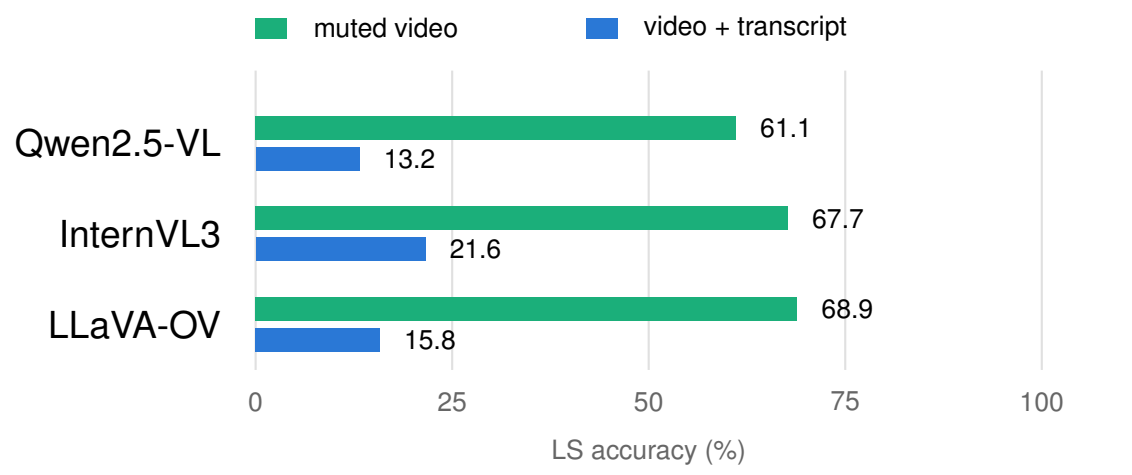
\begin{figure}[t]
\centering
\begin{tikzpicture}[font=\sffamily\scriptsize,x=0.052cm,y=0.6cm]
\foreach \t in {0,25,50,75,100}{
\draw[black!12] (\t,0.2)--(\t,-3.1);
\node[font=\sffamily\tiny,black!60] at (\t,-3.45){\t};}
\foreach \y/\name/\muted/\full in {0.6/{Qwen2.5-VL}/61.1/13.2,1.6/{InternVL3}/67.7/21.6,2.6/{LLaVA-OV}/68.9/15.8}{
\node[anchor=east] at (-2,-\y){\name};
\fill[cvis] (0,-\y+0.04) rectangle (\muted,-\y+0.3);
\fill[ctext] (0,-\y-0.3) rectangle (\full,-\y-0.04);
\node[anchor=west,font=\sffamily\tiny] at (\muted+0.7,-\y+0.17){\muted};
\node[anchor=west,font=\sffamily\tiny] at (\full+0.7,-\y-0.17){\full};}
\fill[cvis] (0,0.55) rectangle (4,0.78);
\node[anchor=west,font=\sffamily\tiny] at (5,0.67){muted video};
\fill[ctext] (42,0.55) rectangle (46,0.78);
\node[anchor=west,font=\sffamily\tiny] at (47,0.67){video + transcript};
\node[font=\sffamily\tiny,black!60] at (50,-4.0){LS accuracy (\%)};
\end{tikzpicture}
\vspace{-4pt}
\caption{\textbf{Adding the transcript} on the same 620 LS clips. Accuracy falls by 47.9, 46.1 and 53.1 points for Qwen2.5-VL, InternVL3 and LLaVA-OV, respectively.}
\label{fig:words}
\vspace{-6pt}
\end{figure}

\para{Zero-shot MLLMs.}
\Cref{tab:s-mrbh} lists every model and view. On the reported view, MLLMs solve 5.5--32.7\% of the LS clips and
89.6--99.3\% of the NS clips; no model reaches chance on LS, and the best LS accuracy belongs to a model that hears
the clip without a transcript (Qwen2-Audio-7B). The muted-video view shows that the evidence is visible: the three
vision--language models solve 61.1--68.9\% of the LS clips from the muted video and 13.2--21.6\% once the transcript is
added (\cref{fig:words}), while the same muted view is the unreliable one on NS clips (40.7--60.7\%). A larger model
of the same family gains little (Qwen2.5-Omni-7B \vs 3B: 28.7\% \vs 25.8\%).

\para{Text inversion.}
Reversing the valid answers of the text-only model solves 94.4\% of the LS clips but 0.7\% of the NS clips. A high LS
accuracy can therefore be produced without any non-verbal input, which is why every LS result is reported with its NS
counterpart and with paired views of the same model.

\para{Fusion models.}
A fusion model is trained on the corpus its clips come from, which gives it two advantages that a zero-shot reader does not have: the label prior of that corpus and its in-domain words. The labels of the LS clips are 61.9\% negative (all MUStARD++ clips and 79\% of the MELD clips), so the constant answer ``negative'' already solves 61.9\% of them and 83.6\% of the NS clips, and the text-only view of LLM-Fusion solves 44.0\% of the LS clips (56\% on IEMOCAP, whose scripted scenes recur across sessions) where the text-only LLM solves 5.5\%. The pooled LS accuracy of the re-run fusion baselines (52.7--60.7\%) is therefore not comparable with that of the MLLMs; \cref{tab:s-mrbh-source} separates the sources. Within the family, CoVeR reaches 61.7\% LS and 93.6\% NS accuracy against 58.8\% and 93.1\% for LLM-Fusion; robust mode trades NS for LS accuracy (63.0\% / 84.8\%). Paired views give evidence that does not depend on the label prior: of the LS clips that a model solves from audio and video alone, adding the text makes 32.2\% wrong for LLM-Fusion and 30.1\% for CoVeR. The inference-time comparator in \cref{tab:s-mrbh} adapts EmoMM-style conflict-triggered steering~\cite{emomm} to LLM-Fusion (it does not reproduce CHASE's head-level attention intervention): when the fused and the text-view states differ by more than a threshold, the fusion-token states of the last quarter of the layers are pushed along their difference; the threshold triggers on 15\% of the validation samples and the strength ($\alpha\in\{0.1,0.3,1\}$) is chosen by validation Acc2. It reaches 61.2\% LS and 91.2\% NS accuracy.

\begin{table*}[!t]
\centering
\caption{\textbf{Results per availability pattern} (Acc2 in \%, MAE, Corr) of the runs in \cref{tab:sota}; mean over 3 seeds.}
\label{tab:s-full}
\footnotesize
\setlength{\tabcolsep}{3pt}
\begin{tabular}{@{}ll ccc ccc ccc ccc@{}}
\toprule
\multirow[c]{2}{*}[-2.4pt]{Pattern} &\multirow[c]{2}{*}[-2.4pt]{Method} & \multicolumn{3}{c}{CMU-MOSI} & \multicolumn{3}{c}{CMU-MOSEI} & \multicolumn{3}{c}{CH-SIMS} & \multicolumn{3}{c}{CH-SIMSv2}\\
\cmidrule(lr){3-5}\cmidrule(lr){6-8}\cmidrule(lr){9-11}\cmidrule(l){12-14}
 & & Acc2\up & MAE\dn & Corr\up & Acc2\up & MAE\dn & Corr\up & Acc2\up & MAE\dn & Corr\up & Acc2\up & MAE\dn & Corr\up\\
\midrule
\multirow{2}{*}{$\{t,a,v\}$} & LLM-Fusion & 80.2&0.899&0.689 & 84.2&0.557&0.753 & 79.1&0.424&0.625 & 83.9&0.294&0.740\\
 & CoVeR & 79.4&0.899&0.708 & 84.8&0.547&0.759 & 83.8&0.403&0.668 & 85.4&0.281&0.754\\
\multirow{2}{*}{$\{t,a\}$} & LLM-Fusion & 79.6&0.899&0.686 & 84.2&0.566&0.746 & 76.8&0.456&0.551 & 78.5&0.342&0.645\\
 & CoVeR & 79.9&0.887&0.714 & 85.2&0.549&0.757 & 80.2&0.439&0.577 & 79.4&0.332&0.654\\
\multirow{2}{*}{$\{t,v\}$} & LLM-Fusion & 79.1&0.908&0.685 & 84.3&0.557&0.753 & 78.8&0.430&0.616 & 83.6&0.302&0.729\\
 & CoVeR & 78.5&0.915&0.704 & 85.1&0.544&0.761 & 83.5&0.407&0.668 & 85.0&0.295&0.742\\
\multirow{2}{*}{$\{a,v\}$} & LLM-Fusion & 53.9&1.400&0.193 & 67.3&0.820&0.310 & 70.4&0.513&0.439 & 72.9&0.388&0.546\\
 & CoVeR & 56.4&1.385&0.228 & 67.7&0.810&0.337 & 74.5&0.503&0.469 & 74.7&0.374&0.583\\
\multirow{2}{*}{$\{t\}$} & LLM-Fusion & 79.1&0.907&0.682 & 84.3&0.566&0.745 & 77.0&0.462&0.543 & 78.7&0.351&0.631\\
 & CoVeR & 78.7&0.931&0.705 & 84.9&0.548&0.755 & 79.3&0.447&0.574 & 79.9&0.347&0.636\\
\multirow{2}{*}{$\{a\}$} & LLM-Fusion & 50.9&1.422&0.174 & 65.0&0.839&0.237 & 61.0&0.576&0.173 & 58.3&0.488&0.198\\
 & CoVeR & 55.7&1.390&0.235 & 65.9&0.815&0.268 & 63.8&0.569&0.170 & 59.7&0.477&0.260\\
\multirow{2}{*}{$\{v\}$} & LLM-Fusion & 51.5&1.456&0.152 & 64.2&0.829&0.263 & 73.4&0.530&0.444 & 72.7&0.399&0.521\\
 & CoVeR & 54.9&1.424&0.188 & 65.3&0.802&0.289 & 75.3&0.514&0.482 & 74.3&0.391&0.547\\
\midrule
\multirow{2}{*}{Avg} & LLM-Fusion & 67.7&1.127&0.504 & 76.2&0.676&0.587 & 73.8&0.484&0.496 & 75.5&0.366&0.593\\
 & CoVeR & 69.1&1.119&0.536 & 77.0&0.659&0.603 & 77.2&0.469&0.530 & 76.9&0.357&0.615\\
\multirow{2}{*}{Rate} & LLM-Fusion & 69.3&1.092&0.543 & 77.7&0.659&0.618 & 74.9&0.476&0.521 & 77.0&0.352&0.624\\
 & CoVeR & 70.3&1.086&0.567 & 78.7&0.637&0.639 & 77.8&0.456&0.567 & 78.3&0.343&0.641\\
\bottomrule
\end{tabular}
\end{table*}

\section{Analyses referenced in the main paper}
\label{sec:s-diag}

\para{Scaling the LLM backbone (\cref{sec:scaling}).}
\Cref{tab:scaling} lists \base with five decoder-only LLMs from 0.5B to 1.7B parameters on CH-SIMS, everything else
fixed. The correlation stays at 48.9--49.9 and the conflict accuracy varies between 35.5\% and 46.8\% without a
consistent trend in size: it falls from 41.9\% to 35.5\% from Qwen2.5-0.5B to 1.5B and rises from 38.7\% to 46.8\% from
Qwen3-0.6B to 1.7B. Every backbone still follows the text on 53--65\% of the conflict cases, although it solves
63--73\% of them once the text is removed. Over three seeds on CMU-MOSI, CH-SIMS and CH-SIMSv2, replacing Qwen3-0.6B by
Qwen3-1.7B changes the average correlation of \base by $-$1.3, $+$0.2 and $-$2.1 points and the conflict accuracy by
$+$7.4, $+$7.5 and $-$5.5 points, leaving \shorterr at 70.4\%, 53.8\% and 46.1\%. The zero-shot MLLMs tell the same
story: Qwen2.5-Omni-7B solves 178 of the 620 LS clips against 160 for its 3B sibling (\cref{fig:results} of
the main paper).

\begin{table}[H]
\centering
\caption{\textbf{Does scaling up solve the problem?} LLM-Fusion with five LLM backbones on CH-SIMS (all modalities observed):
Avg Corr ($\times100$) over the seven patterns, Acc2 on MR-Bench-A consensus and conflict cases, ShortErr, and conflict
Acc2 of the same model with the text removed (\%). Mean over 2 seeds (3 for Qwen3-0.6B).
Pink, orange and yellow mark the best, second-best and third-best distinct values, respectively; ties share a color.}
\label{tab:scaling}
\footnotesize
\setlength{\tabcolsep}{4pt}
\renewcommand{\arraystretch}{1.05}
\begin{tabular}{@{}lc ccccc@{}}
\toprule
Backbone & Params & Corr\up & Cons.\up & Conf.\up & SE\dn & w/o $t$\up \\
\midrule
Qwen2.5-0.5B~\cite{qwen25}
& 0.5B
& \cellcolor{best}49.9
& 89.4
& \cellcolor{second}41.9
& \cellcolor{second}58.1
& \cellcolor{best}72.6 \\
Qwen3-0.6B~\cite{qwen3}
& 0.6B
& \cellcolor{second}49.6
& \cellcolor{third}89.6
& \cellcolor{third}38.7
& \cellcolor{third}61.3
& \cellcolor{second}68.8 \\
Qwen2.5-1.5B~\cite{qwen25}
& 1.5B
& \cellcolor{third}49.4
& \cellcolor{second}90.0
& 35.5
& 64.5
& 62.9 \\
Qwen3-1.7B~\cite{qwen3}
& 1.7B
& \cellcolor{best}49.9
& 88.2
& \cellcolor{best}46.8
& \cellcolor{best}53.2
& \cellcolor{third}67.7 \\
SmolLM2-1.7B~\cite{smollm2}
& 1.7B
& 48.9
& \cellcolor{best}90.3
& \cellcolor{second}41.9
& \cellcolor{second}58.1
& 66.1 \\
\bottomrule
\end{tabular}
\vspace{-6pt}
\end{table}

\para{Test-time override sweep (\cref{sec:override}).}
\Cref{tab:override} varies the test-time margin $\tau_{\mathrm{test}}$ of the override applied to the trained \base.
A smaller margin lets the views overrule the text more often and raises the conflict accuracy, but the consensus
accuracy and the correlation fall on every dataset. The CMU-MOSI/CH-SIMSv2 rows at $\tau_{\mathrm{test}}=0.3$
are also reported in \cref{tab:reliability}; the CH-SIMS sweep retains its earlier reference checkpoint series.

\begin{table}[H]
\centering
\caption{\textbf{Test-time override trades consensus for conflict accuracy.} LLM-Fusion with disagreeing views overriding the
fused decision (\cref{eq:gate,eq:mix}) at test-time margin $\tau_{\mathrm{test}}$ ($-$: no override). Avg Corr
($\times100$) and MR-Bench-A Acc2 (\%) on consensus (Cons.) and conflict (Conf.) cases; all rows from the same evaluation of
LLM-Fusion with its views. Mean over 5 seeds. This is the earlier margin-sweep checkpoint series, separate from the updated CH-SIMS ablation.}
\label{tab:override}
\scriptsize
\setlength{\tabcolsep}{2.3pt}
\begin{tabular}{@{}c ccc ccc ccc@{}}
\toprule
\multirow[c]{2}{*}[-2.4pt]{$\tau_{\mathrm{test}}$} & \multicolumn{3}{c}{CMU-MOSI} & \multicolumn{3}{c}{CH-SIMS} & \multicolumn{3}{c}{CH-SIMSv2}\\
\cmidrule(lr){2-4}\cmidrule(lr){5-7}\cmidrule(l){8-10}
 & Corr\up & Cons.\up & Conf.\up & Corr\up & Cons.\up & Conf.\up & Corr\up & Cons.\up & Conf.\up\\
\midrule
$-$ &  50.7&94.1&21.7 & 48.8&89.2&41.3 & 59.2&93.7&58.6 \\
0.5 &  50.3&93.8&23.3 & 45.9&84.4&51.6 & 58.1&92.3&63.6 \\
0.4 &  49.3&93.5&26.7 & 44.4&81.7&59.4 & 57.0&90.5&67.4 \\
0.3 &  47.8&92.5&33.9 & 42.9&78.5&62.6 & 55.3&86.8&72.3 \\
0.2 &  44.2&86.4&48.9 & 41.5&73.8&67.1 & 52.9&78.8&79.5 \\
0.1 &  39.0&72.3&66.7 & 40.8&69.8&69.0 & 50.0&67.3&81.1 \\
\bottomrule
\end{tabular}
\vspace{-6pt}
\end{table}

\para{Corrections compared in \cref{tab:reliability} (\cref{sec:cover}).}
The \emph{learned reliability gate} runs the counterfactual views
(text-, \mbox{audio-,} visual-only and content-free) first, without gradient. For each view, a small gate reads the fused
readout, the readout of the view and the predictions of the views and outputs a strength in $(0,1)$; the fused readout is
pushed away from the views in proportion to these strengths, which are budgeted to sum to at most one. The gates of the
single-modality views are trained to fire on unreliable views: the target is 1 if the judgment of the modality disagrees
with the label and 0 otherwise, where the judgment is the human unimodal label (CH-SIMSv2) or the prediction of the view
itself (CMU-MOSI). The views are needed at test time as well. \emph{Conflict re-weighting} trains \base with the task
loss weighted by 5 on the conflict cases of the training split (\cref{sec:conflict}). \emph{Unimodal multi-task} is \ours
without the gate (\cref{eq:uni}).

\section{Permanent missingness}
\label{sec:s-fixed}

\begin{table}[H]
\centering
\caption{\textbf{Dynamic masking \vs permanent missingness.} Avg Corr ($\times100$) over the seven test patterns. Dyn.: the main protocol (\cref{tab:sota}); Perm.: every training and validation sample keeps one availability pattern for the whole run. ${}^{\dagger}$: the losses of these methods target the features of missing modalities, which permanent missingness does not provide; they receive them as an oracle. No complete: only the six incomplete patterns are drawn. Mean over 3 seeds; best permanent result in bold.}
\label{tab:s-fixed}
\scriptsize
\setlength{\tabcolsep}{1.9pt}
\begin{tabular}{@{}l cc cc cc cc@{}}
\toprule
\multirow[c]{2}{*}[-2.4pt]{Method} & \multicolumn{2}{c}{CMU-MOSI} & \multicolumn{2}{c}{CMU-MOSEI} & \multicolumn{2}{c}{CH-SIMS} & \multicolumn{2}{c}{CH-SIMSv2}\\
\cmidrule(lr){2-3}\cmidrule(lr){4-5}\cmidrule(lr){6-7}\cmidrule(l){8-9}
 & Dyn. & Perm. & Dyn. & Perm. & Dyn. & Perm. & Dyn. & Perm.\\
\midrule
MulT~\cite{mult} & 52.1 & 47.9 & 53.2 & 51.2 & 41.4 & 44.8 & 53.4 & 52.1\\
MISA~\cite{misa} & 47.9 & 44.5 & 54.9 & 52.3 & 46.3 & 45.8 & 55.0 & 54.9\\
MoMKE~\cite{momke} & 47.4 & 42.2 & 57.4 & 54.7 & 44.6 & 42.2 & 57.9 & 54.5\\
MPLMM~\cite{mplmm} & 52.0 & 46.2 & 56.7 & 54.4 & 50.0 & 43.2 & 58.1 & 56.6\\
\midrule
MMIN~\cite{mmin}$^{\dagger}$ & 50.4 & 46.9 & 57.4 & 55.7 & 50.8 & 45.7 & 57.7 & 57.1\\
TFR-Net~\cite{tfrnet}$^{\dagger}$ & 50.1 & 44.5 & 51.5 & 51.7 & 46.2 & 43.3 & 56.5 & 54.5\\
IMDer~\cite{imder}$^{\dagger}$ & 50.6 & 45.2 & 56.7 & 54.4 & 47.8 & 45.5 & 58.1 & 54.5\\
DiCMoR~\cite{dicmor}$^{\dagger}$ & 51.1 & 45.1 & 56.4 & 53.8 & 52.2 & 46.0 & 59.3 & 54.4\\
LNLN~\cite{lnln}$^{\dagger}$ & 47.7 & 47.2 & 54.3 & 53.3 & 52.6 & 49.2 & 57.2 & 55.3\\
\midrule
LLM-Fusion & 50.4 & 47.8 & 58.7 & 56.0 & 49.6 & 46.6 & 59.3 & 57.9\\
\rowcolor{oursrow}CoVeR (ours) & 53.6 & \textbf{49.5} & 60.3 & \textbf{58.8} & 53.0 & \textbf{50.0} & 61.5 & \textbf{59.8}\\
\midrule
LLM-Fusion, no complete & -- & 50.5 & -- & 57.4 & -- & 46.3 & -- & 58.1\\
CoVeR, no complete & -- & 52.9 & -- & 59.1 & -- & 48.8 & -- & 58.5\\
\bottomrule
\end{tabular}
\end{table}

\para{Protocol.}
In the main protocol the availability pattern of a training sample is re-drawn at every visit, so over a run each of
its modalities is eventually observed. Under \emph{permanent missingness} every training and validation sample is
assigned one of the seven non-empty patterns once, before training (balanced over the patterns and independent of the
training seed), and keeps it for the whole run. The features of its unobserved modalities are zeroed when the data are
loaded, so no component can read them: not the model, not the unimodal judges or their targets (on CMU-MOSI/MOSEI the
expert of a modality is trained and applied only on the samples that observe it), and not model selection, which scores
the validation split under its own permanent patterns. A modality of a training sample is thus observed with
probability $4/7$ and never otherwise, and one sample in seven is complete; a stricter variant draws only the six
incomplete patterns, so that no training or validation sample is ever complete. The test protocol is unchanged (seven
patterns on the complete test split) and every method keeps its recipe of \cref{tab:sota}. Methods whose losses target
the features of missing modalities (T-Miss \xmark{} in \cref{tab:sota}) cannot be trained as designed in this setting;
we grant them these targets as an oracle (${}^{\dagger}$ in \cref{tab:s-fixed}).

\para{Results.}
Permanent missingness lowers average correlation across datasets for every method (\cref{tab:s-fixed}): averaged over the four datasets, the nine IML methods lose 2.8 Avg Corr points, LLM-Fusion 2.4 and CoVeR about 2.6, although five of the nine keep oracle access to the features they can no longer observe. CoVeR remains the best method on all four datasets: 49.5/58.8/50.0/59.8 on CMU-MOSI/MOSEI/CH-SIMS/CH-SIMSv2, against 47.8/56.0/46.6/57.9 for LLM-Fusion and at most 47.9/55.7/49.2/57.1 for the re-run methods (MulT, MMIN, LNLN). Its margin over LLM-Fusion is +1.7/+2.8/+3.4/+1.9 points, compared with approximately +3.2/+1.6/+3.4/+2.2 under dynamic masking, where the judges see every modality of every sample over a run.
With no complete training or validation sample at all, CoVeR reaches 52.9/59.1/48.8/58.5 against 50.5/57.4/46.3/58.1 for LLM-Fusion on CMU-MOSI/CMU-MOSEI/CH-SIMS/CH-SIMSv2.

\section{Additional results}
\label{sec:s-results}

\Cref{tab:iemocap} compares \ours with the nine re-run methods on IEMOCAP, and \cref{tab:s-full} lists Acc2, MAE and
Corr of \base and \ours for every availability pattern, the average over patterns and the average over random missing
rates on the four MMSA datasets.

\newpage
\begin{table}[H]
\centering
\caption{\textbf{IEMOCAP} (features extracted by us, supplementary material): mean over the seven availability patterns,
Acc2 and F1 in \%. LLM-Fusion: the backbone of CoVeR without judges and gate. Mean over 3 seeds.
Pink, orange and yellow mark the best, second-best and third-best distinct values, respectively; ties share a color.}
\label{tab:iemocap}
\footnotesize
\setlength{\tabcolsep}{4pt}
\begin{tabular}{@{}llc cccc@{}}
\toprule
Method & Venue & T-Miss & Acc2\up & F1\up & MAE\dn & Corr\up\\
\midrule
MulT~\cite{mult} & ACL'19 & \cmark & 77.9 & 75.8 & 0.616 & 0.556 \\
MISA~\cite{misa} & MM'20 & \cmark & 80.2 & 77.9 & 0.599 & 0.602 \\
MMIN~\cite{mmin} & ACL'21 & \xmark & \cellcolor{third}80.9 & \cellcolor{third}80.6 & 0.626 & \cellcolor{third}0.627 \\
TFR-Net~\cite{tfrnet} & MM'21 & \xmark & 77.9 & 76.7 & 0.612 & 0.562 \\
IMDer~\cite{imder} & NeurIPS'23 & \xmark & 79.2 & 78.9 & \cellcolor{third}0.579 & 0.606 \\
DiCMoR~\cite{dicmor} & ICCV'23 & \xmark & 80.1 & 79.6 & 0.580 & 0.615 \\
MoMKE~\cite{momke} & MM'24 & \cmark & 80.2 & 79.9 & 0.585 & 0.615 \\
LNLN~\cite{lnln} & NeurIPS'24 & \xmark & 78.0 & 77.7 & 0.608 & 0.594 \\
MPLMM~\cite{mplmm} & ACL'24 & \cmark & 78.8 & 78.7 & 0.588 & 0.604 \\
\midrule
LLM-Fusion & -- & \cmark & \cellcolor{second}82.3 & \cellcolor{second}82.4 & \cellcolor{second}0.538 & \cellcolor{second}0.671 \\
CoVeR (ours) & -- & \cmark & \cellcolor{best}82.6 & \cellcolor{best}82.6 & \cellcolor{best}0.534 & \cellcolor{best}0.681 \\
\multicolumn{3}{@{}l}{\quad $\Delta$ \vs LLM-Fusion} & +0.3 & +0.2 & -0.004 & +0.010 \\
\bottomrule
\end{tabular}
\end{table}

\end{document}